%% file: manuscript.tex
\documentclass[onecolumn, 10pt, final, compsoc]{IEEEtran}
\IEEEoverridecommandlockouts
\usepackage{cite}
\usepackage{amsmath,amssymb,amsfonts, amsthm}
\usepackage{algorithmicx}
\usepackage{algpseudocode}
\usepackage{algorithm}
\usepackage{graphicx}
\usepackage{svg}
\usepackage{textcomp}
\usepackage{mathtools}
\usepackage{booktabs}
\usepackage{multirow}
\usepackage{longtable}
\usepackage{soul} %
\usepackage[colorlinks]{hyperref}
\usepackage[acronym,nonumberlist]{glossaries}
\usepackage[capitalize]{cleveref}
\usepackage{subcaption}
\usepackage[section]{placeins}
\usepackage{lscape} %
\usepackage{verbatim}
\usepackage{listings}
\usepackage{array}
\usepackage{dirtytalk}
\usepackage[utf8]{inputenc}
\usepackage{setspace}
\usepackage{booktabs} %
\usepackage{siunitx} %
\usepackage{tikz}
\usepackage{makecell}
\usepackage{caption} %
\usepackage{wrapfig}
\usepackage{dsfont}

\usepackage{tablefootnote} %

\crefname{listing}{Code Block}{Code Blocks}

\definecolor{darkgreen}{rgb}{0.0, 0.5, 0.0}
\definecolor{purple}{rgb}{0.5, 0.0, 0.5}
\definecolor{backcolour}{rgb}{0.95,0.95,0.92}
\definecolor{codegray}{rgb}{0.5,0.5,0.5}
\lstdefinestyle{python}
{
  language=Python,                
  basicstyle=\normalsize\ttfamily,
  numbers=left,    %
  backgroundcolor=\color{backcolour},  %
  frame=single,                   %
  numberstyle=\tiny\color{codegray},
  numbers=left,
  rulecolor=\color{codegray},     %
  tabsize=2,      %
  breaklines=true,   %
  breakatwhitespace=false, %
  keywordstyle=\color{blue},      %
  commentstyle=\color{darkgreen},     %
  stringstyle=\color{purple} %
}

\makeglossaries
\usepackage{framed,enumitem}

\hypersetup{
    colorlinks  = true,
    linkcolor   = blue,   %
    anchorcolor = black,
    citecolor   = blue,   %
    filecolor   = cyan,
    menucolor   = red,
    runcolor    = cyan,
    urlcolor    = blue    %
}
\def\BibTeX{{\rm B\kern-.05em{\sc i\kern-.025em b}\kern-.08em
    T\kern-.1667em\lower.7ex\hbox{E}\kern-.125emX}}

\newcommand{\anonymous}[1]{\textcolor{blue}{Anonymous (withheld for blind peer-review)}} %
\newcommand{\etal}{\textit{et al. }}

\theoremstyle{definition}

\newcommand{\overbar}[1]{\mkern 1.5mu\overline{\mkern-1.5mu#1\mkern-1.5mu}\mkern 1.5mu}
\DeclareMathOperator{\EX}{\mathbb{E}}

\newacronym{nserc}{NSERC}{Natural Sciences and Engineering Research Council of Canada}
\newacronym{oce}{OCE}{Ontario Center of Excellence}
\newacronym{ppe}{PPE}{Personal Protective Equipment}
\newacronym{osh}{OSH}{Occupational Safety and Health}
\newacronym{osha}{OSHA}{Occupational Safety and Health Administration}
\newacronym{dl}{DL}{Deep Learning}
\newacronym{od}{OD}{Object Detection}
\newacronym{cv}{CV}{Computer Vision}
\newacronym{cnn}{CNN}{Convolutional Neural Network}
\newacronym{yolo}{YOLO}{You Only Look Once}
\newacronym{covid}{COVID-19}{Coronavirus Disease}
\newacronym{gpu}{GPU}{Graphical Processing Unit}
\newacronym{rcnn}{R-CNN}{Region Proposal Convolutional Neural Network}
\newacronym{fpn}{FPN}{Feature Pyramid Network}
\newacronym{cspnet}{CSPNet}{Cross-Stage Partial Connections}
\newacronym{panet}{PANet}{Path Aggregation Network}
\newacronym{sam}{SAM}{Spatial Attention Module}
\newacronym{ciou}{CIoU}{Complete Intersection Over Union}
\newacronym{silu}{SiLU}{Sigmoid-weighted Linear Unit}
\newacronym{elan}{ELAN}{Efficient Long-Range Attention Network}
\newacronym{gelan}{GELAN}{Generalized Efficient Long-Range Attention Network}
\newacronym{repconvn}{RepConvN}{Reparameterized Convolutional}
\newacronym{dfl}{DFL}{Distribution Focal Loss}
\newacronym{lfyolo}{LF-YOLO}{Lighter and Faster YOLO}
\newacronym{rmf}{RMF}{Reinforced Multiscale Feature}
\newacronym{attyolo}{ATT-YOLO}{Attention YOLO}
\newacronym{senet}{SENet}{Squeeze and Excitation Network}
\newacronym{yoloimf}{YOLO-IMF}{YOLO for Industrial Manufacturing Field}
\newacronym{mscoco}{MS-COCO}{Microsoft Common Objects in Context}
\newacronym{rtx}{RTX}{Ray Tracing Texel eXtreme}
\newacronym{p}{P}{Precision}
\newacronym{r}{R}{Recall}
\newacronym{tp}{TP}{True Positive}
\newacronym{fp}{FP}{False Positive}
\newacronym{tn}{TN}{True Negative}
\newacronym{fn}{FN}{False Negative}
\newacronym{map}{mAP}{Mean Average Precision}
\newacronym{iou}{IoU}{Intersection over Union}
\newacronym{flops}{FLOPS}{Floating-Point Operations Per Second}
\newacronym{id}{ID}{Identifier}
\newacronym{deepsort}{DeepSort}{Deep Simple Object Tracking}
\newacronym{kf}{KF}{Kalman Filtering}
\newacronym{mes}{MES}{Manufacturing Execution System}
\newacronym{lmm}{LMM}{Large Multimodal Model}
\newacronym{ssim}{SSIM}{Structural Similarity Index Measure}
\newacronym{tsne}{t-SNE}{T-distributed Stochastic Neighbor Embedding}
\newacronym{fps}{fps}{Frames Per Second}
\newacronym{ml}{ML}{Machine Learning}
\newacronym{iot}{IoT}{Internet of Thing}
\newacronym{roi}{RoI}{Region of Interest}
\newacronym{ssd}{SSD}{Single Stage Detector}
\newacronym{nms}{NMS}{Non-Max-Suppression}
\newacronym{pgi}{PGI}{Programmable Gradient Information}
\newacronym{ai}{AI}{Artificial Intelligence}
\newacronym{nn}{NN}{Neural Network}
\newacronym{ood}{OOD}{Out-of-Distribution}
\newacronym{da}{DA}{Data Augmentation}
\newacronym{iid}{IID}{Independant and Identically Distributed}
\newacronym{vrm}{VRM}{Vicinal Risk Minimization}
\newacronym{erm}{ERM}{Empirical Risk Minimization}
\newacronym{jsd}{JSD}{Jensen–Shannon Divergence}
\newacronym{pgd}{PGD}{Projected Gradient Descent}
\newacronym{ce}{CE}{Corruption Error}
\newacronym{bce}{BCE}{Binary Cross Entropy}

\begin{document}
\title{LayerMix: Enhanced Data Augmentation through Fractal Integration for Robust Deep Learning
\thanks{This study is supported by IFIVEO CANADA INC., Mitacs through IT16094, and the University of Windsor, Canada.}
}

\author{
    \IEEEauthorblockN{Hafiz Mughees Ahmad\IEEEauthorrefmark{2} ~ Dario Morle \IEEEauthorrefmark{3} ~ Afshin Rahimi\IEEEauthorrefmark{2}} \\
    \IEEEauthorblockA{\IEEEauthorrefmark{2}University of Windsor, Canada and     \IEEEauthorrefmark{3}IFIVEO CANADA INC.\\
    \{ahmad54, arahimi\}@uwindsor.ca~  dario@ifiveo.com
}}

\maketitle
\begin{abstract}
Deep learning models have demonstrated remarkable performance across various computer vision tasks, yet their vulnerability to distribution shifts remains a critical challenge. Despite sophisticated neural network architectures, existing models often struggle to maintain consistent performance when confronted with \acrfull{ood} samples, including natural corruptions, adversarial perturbations, and anomalous patterns.
We introduce LayerMix, an innovative data augmentation approach that systematically enhances model robustness through structured fractal-based image synthesis. By meticulously integrating structural complexity into training datasets, our method generates semantically consistent synthetic samples that significantly improve neural network generalization capabilities. Unlike traditional augmentation techniques that rely on random transformations, LayerMix employs a structured mixing pipeline that preserves original image semantics while introducing controlled variability.
Extensive experiments across multiple benchmark datasets, including CIFAR-10, CIFAR-100, ImageNet-200, and ImageNet-1K demonstrate LayerMix's superior performance in classification accuracy and substantially enhances critical \acrfull{ml} safety metrics, including resilience to natural image corruptions, robustness against adversarial attacks, improved model calibration and enhanced prediction consistency. LayerMix represents a significant advancement toward developing more reliable and adaptable artificial intelligence systems by addressing the fundamental challenges of deep learning generalization. The code is available at \url{https://github.com/ahmadmughees/layermix}.
\end{abstract}

\begin{IEEEkeywords}
Classification, Data Augmentation, Fractals, Robustness, Corruption, Adversarial Attack. 
\end{IEEEkeywords}

\newcommand{\glsentrysymbolorhyphen}[1]{%
  \glsletentryfield{\currentsymbol}{#1}{symbol}%
  \ifx\currentsymbol\empty%
    $-$%
  \else%
    \currentsymbol%
  \fi%
}

\newglossarystyle{mynomenclstyle}{
  \setglossarystyle{long} %
  \renewenvironment{theglossary}{
    \begin{longtable}{p{0.1\textwidth}p{0.65\textwidth}>{\centering\arraybackslash}p{0.1\textwidth}} %
  }{
    \end{longtable}
  }
  \renewcommand*{\glossaryheader}{
    \bfseries Term & \bfseries Description & \bfseries Unit \\
    \hline
    \endhead
  }
  \renewcommand*{\glossentry}[2]{
    \glstarget{##1}{\glossentryname{##1}} & %
    \glossentrydesc{##1} & %
    \glsentrysymbolorhyphen{##1}\\ %
  }
}

\newglossarystyle{myacrostyle}{
  \setglossarystyle{long} %
  \renewenvironment{theglossary}{
    \begin{longtable}{p{0.1\textwidth}p{0.75\textwidth}} %
  }{
    \end{longtable}
  }
  \renewcommand*{\glossaryheader}{
    \bfseries Term & \bfseries Description \\ %
    \hline
    \endhead
  }
  \renewcommand*{\glossentry}[2]{
    {\glstarget{##1}{\glossentryname{##1}}} & %
    {\glossentrydesc{##1}}\\
    
  }
}

\section{Introduction}
\acrfull{dl} based models have proven highly effective \cite{cubukAutoAugmentLearningAugmentation2019, cubukRandAugmentPracticalAutomated2019, Zhong2017RandomED} in training \acrfull{cv} tasks ~\cite{Wang2018ZeroShotIC,Zhai2021LiTZT,Gu2021OpenvocabularyOD, Maturana2015VoxNetA3,Yang2018PIXORR3,Mao2021VoxelTF, Wang2017NormFaceLH,Wang2018CosFaceLM}, including but not limited to image classification ~\cite{heDeepResidualLearning2016, zagoruykoWideResidualNetworks2016, szegedyInceptionv4InceptionresnetImpact2017, tanEfficientNetRethinkingModel2019}, object detection~\cite{wangYOLOv9LearningWhat2024, tanEfficientDetScalableEfficient2020, renFasterRCNNRealTime2017}, and semantic segmentation~\cite{heMaskRCNN2017, ronnebergerUNetConvolutionalNetworks2015, raviSAM2Segment2024}.
While these models perform exceptionally well under ideal conditions where the training and test data follow the same distribution, their robustness is often challenged when faced with \acrfull{ood} samples~\cite{Recht2019DoIC, Croce2020RobustBenchAS, Hendrycks2021UnsolvedPI, hendrycksBenchmarkingNeuralNetwork2018}.
Common \acrshort{ood} scenarios include natural corruptions~\cite{Mintun2021OnIB}, adversarial perturbations~\cite{Metzen2017OnDA}, and anomaly patterns~\cite{Elsayed2020NetworkAD}, highlighting the critical need for models to maintain accuracy across distribution shifts. 
To address this, \acrfull{da} has emerged as a widely adopted strategy, where various transformations are applied to existing images to generate synthetic yet diverse training examples~\cite{Gong2020KeepAugmentAS, Calian2021DefendingAI, Kim2020PuzzleME}. By expanding the diversity of the training dataset, \acrshort{da} can significantly improve the model's robustness against unseen data distribution shifts~\cite{Subbaswamy2021EvaluatingMR}.

\acrshort{da} techniques are traditionally categorized into two main types \cite{wangComprehensiveSurveyData2024, mumuniDataAugmentationComprehensive2022}: 1) Individual augmentations that operate independently on a single data sample to generate new variations. These include both, spatial and affine transformations~\cite{Lim2019FastA, Cubuk2019AutoAugmentLA, cubukRandAugmentPracticalAutomated2019}. 2) Multiple augmentations use multiple samples to synthesize a new sample. It can use the samples from the same dataset or some other data source~\cite{Zhang2018mixupBE, Takahashi2018RICAPRI, Kim2021CoMixupSG}. Recently, a new line of research to create complex images has been proposed \cite{hendrycksPixMixDreamlikePictures2022, huangIPMixLabelpreservingData2024, islamDiffuseMixLabelPreservingData2024, islamGenMixEffectiveData2024} where training images are mixed with Structurally Complex Objects, which is often described in terms of the degree of organization \cite{lloydMeasuresComplexityNonexhaustive2001}. Fractals serve as a classic instance of structurally intricate objects, which were also utilized in the pretraining of image classifiers. \cite{kataokaPretrainingNaturalImages, nakashimaCanVisionTransformers2022}. Fractals are label-preserving approach instead of MixUp~\cite{Zhang2018mixupBE} based approaches where new labels are assigned to synthesized samples, which often results in manifold intrusion \cite{guoMixUpLocallyLinear2019, baenaLocalMixupInterpolation2024}. 

Hendrycks \etal \cite{hendrycksPixMixDreamlikePictures2022} created the labels-preserved complex images using the conic combination of fractals and training images. 
Huang \etal \cite{huangIPMixLabelpreservingData2024} used the multi-level mixing approach of pixel, patch, and image for mixing fractals into training samples to generate unique samples. Both of these approaches improved training accuracy and \acrshort{ml} robustness metrics evaluated on benchmark datasets \cite{hendrycksBenchmarkingNeuralNetwork2018, Hendrycks2019NaturalAE}. These preliminary studies examined the integration of fractals with randomly sampled data in a pipeline characterized by a vast search space, necessitating the use of supplementary training methodologies to enhance optimization. To address this and reduce the dependence on additional training tricks, we have proposed a LayerMix, a structured mixing pipeline that combines unique samples to train a \acrshort{ml} model, achieving improvements across all benchmark metrics. 

The primary contributions of this study are summarized as follows:

\begin{enumerate} 
    \item A comprehensive theoretical framework for investigating optimal strategies for mixing fractals. 
    \item A novel mathematically evaluated mixing pipeline, with each step rigorously analyzed. 
    \item Extensive experimental evaluations demonstrating state-of-the-art performance on the CIFAR-10, CIFAR-100, ImageNet-200, and ImageNet benchmarks, achieving substantial improvements in generalization and adversarial robustness compared to existing methods. 
    \item The full open-sourcing of the framework, enabling evaluation of classification models across all metrics in a unified platform, alongside the release of training runs and experimental meta-data on GitHub. 
\end{enumerate}

\begin{figure}[t]
    \centering
    \includegraphics[width=0.65\textwidth]{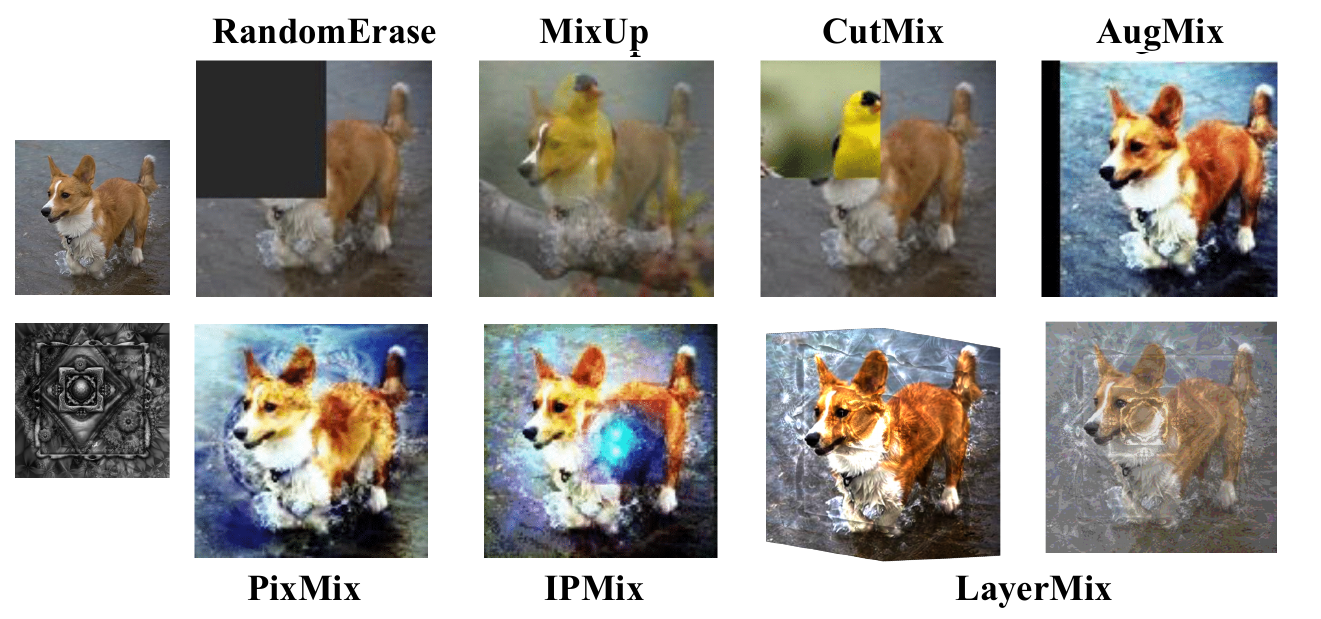}
    \caption{Comparison of different augmentation pipelines.}
    \label{fig:layermix_comp}
\end{figure}

The remainder of this paper is structured as follows: background and literature review are provided in \cref{sec:related_work}. Technical details are provided in \cref{sec:methodology} and discussions on the results of the experiments are provided in \cref{sec:experiments}. Finally, \cref{sec:conclusion} offers concluding remarks and future directions. 
\section{Related Work} \label{sec:related_work}
\begin{center}
\say{Data Augmentation methodology can outperform models trained with 1000x more data - Hendryks \etal~\cite{Hendrycks2020TheMF_imagenet_r}}.     
\end{center}
Data augmentation serves as a pivotal technique in deep learning, aiming to enhance the diversity of training datasets by systematically transforming existing data samples. By creating novel and varied instances, these transformations facilitate better regularization and improve the generalization capacity of models. Since the early days of \acrshort{dl}, including \acrfull{cnn}-based architectures such as LeNet~\cite{lecunGradientbasedLearningApplied1998}, AlexNet~\cite{krizhevskyImagenetClassificationDeep2012}, and ResNet~\cite{heDeepResidualLearning2016}, data augmentation has been a cornerstone in achieving robust performance. It's significance continues with more recent architectures \cite{vaswaniAttentionAllYou2017} such as Vision Transformers~\cite{dosovitskiyImageWorth16x162021, liuSwinTransformerHierarchical2021} and Diffusion Networks~\cite{hoDenoisingDiffusionProbabilistic2020, sohl-dicksteinDeepUnsupervisedLearning2015}. 

Typically, modern \acrshort{dl} models have a representation capacity beyond the datasets they are trained on \cite{double-descent}. However, recent research has shown that there are limits past which sample diversity hinders performance. For example, \cite{guoMixUpLocallyLinear2019} identifies manifold intrusion as a limit where extreme augmentation can cause synthetic labels to collide with labels from the original dataset. To understand these limitations, we standardize the definitions and represent the pipelines in the literature mathematically as a function of data distributions.

\subsection{Affinity and Diversity}
\label{sec:background.affdiv}
Modern data augmentation pipelines aggressively increase diversity, even sampling out of distribution data.  
To understand these pipelines, Lopes \etal~\cite{affinity-diversity} and Yang \etal~\cite{similarity-diversity} identify two inversely proportional metrics for analysis.  In \cite{affinity-diversity}, \textbf{affinity} is defined as the difference in model accuracy between the test-set and the augmented test-set, and final training loss value is used to define \textbf{diversity}.  In \cite{similarity-diversity}, \textbf{similarity} is defined as Wasserstein distances in the feature space of a trained model, and an aggregation of the eigenvalues of per-class feature embeddings is used to define a \textbf{diversity}.

To combine both intuitively for the rest of the article, here we define \textbf{affinity} as a measure for the amount that an augmentation pipeline induces label confusion in the model and  \textbf{diversity} provides a measure for the extent to which the augmentation pipeline extends samples beyond the data manifold.  These metrics will naturally be inversely proportional, requiring augmentation algorithms to optimize both to achieve increased model performance jointly.

\subsection{Augmentation Pipelines}
\label{sec:background.augpipe}

The goal of data augmentation in deep learning is to improve performance by showing models a more diverse set of inputs than would be feasible using raw data directly.
To construct a typical augmentation pipeline, a set of stochastic label-preserving transformations are applied sequentially.
We can represent one of these transforms $f_k$ as a sample $\mathbf{Y}$ drawn from a distribution over the output space, conditioned on the input to the transformation $\mathbf{z}$, i.e., $\mathbf{Y}\sim f_k(\mathbf{y}|\mathbf{z}=\mathbf{Z})$, where $k$ enumerates the different transformations (such as rotation, color jitter, translation).
AutoAugment \cite{Cubuk2019AutoAugmentLA} sought to determine a dataset-specific ordering of transformations through reinforcement learning.  RandAugment \cite{cubukRandAugmentPracticalAutomated2019} eliminated the need for an order selection by randomly choosing a transformation $f_k$ for each stage in the pipeline.  Mathematically, this was equivalent to sampling from $\EX_k\left[f_k(\mathbf{y}|\mathbf{z})\right]$ rather than a particular $f_k(\mathbf{y}|\mathbf{z})$.  This was found to increase sample diversity, ultimately leading to improved performance over AutoAugment.  Additionally, RandAugment provided a simpler implementation since each stage in the pipeline is now programmatically identical.  One implication of this structure is that the joint distribution over the input-output spaces $\mathbf{x}=[\mathbf{y}^T,\mathbf{z}^T]^T$ of the augmentation stage $p(\mathbf{x})=p(\mathbf{y}, \mathbf{z})=p(\mathbf{y}|\mathbf{z})p(\mathbf{z})=p(\mathbf{z})\EX_k\left[f_k(\mathbf{y}|\mathbf{z})\right]$ was \acrfull{iid} across all stages in the pipeline.

Outside of augmentation pipeline improvements, additional methods for increasing the efficacy of augmentation were proposed as well.  MixUp \cite{Zhang2018mixupBE} generated augmented samples by linearly interpolating between two randomly selected images and their corresponding labels, encouraging smoother decision boundaries for classification tasks.  Manifold-MixUp \cite{Verma2018ManifoldMB} performs similar interpolations within the hidden layers of a \acrshort{nn}, leading to improved accuracy by enforcing smooth transitions between feature representations.  Both of these methods utilized \acrfull{vrm} \cite{chapelle2000vicinal} to increase the diversity of samples.

The \acrshort{iid} augmentation stages proposed by RandAugment \cite{cubukRandAugmentPracticalAutomated2019} were combined with the blending stages of MixUp \cite{Zhang2018mixupBE} in AugMix \cite{hendrycks*AugMixSimpleData2019} to further increase diversity, resulting in improved model performance.  This improvement was especially notable on robustness benchmarks.
AugMix followed MixUp regarding the implementation of the blending stage.  An arithmetic mean of images was used, weighted by a random convex combination over the images.  The images used for blending consisted primarily of augmentation stages applied to the original image and blends thereof.
AugMix also included a new term in the loss function named \acrfull{jsd} consistency to increase model performance.  This new term ensured that the learned representations of an augmented image did not diverge significantly from the representation of the original image.  This is a form of affinity as defined in section \ref{sec:background.affdiv}.

PixMix \cite{hendrycksPixMixDreamlikePictures2022} built on AugMix \cite{hendrycks*AugMixSimpleData2019} through the introduction of fractals for blending, pipeline structure improvements, and increasing the resultant diversity from blending stages.  In AugMix, each blending stage would use a weighted arithmetic mean of images.  PixMix extended this by considering a mixture of blending methods, similar to the extension RandAugment used on AutoAugment.  Mathematically, if a blending method is described as sampling an image $\mathbf{Y}$ from a distribution $g_k$ conditioned on two input images $\mathbf{Z_0}$ and $\mathbf{Z_1}$: $\mathbf{Y}\sim g_k(\mathbf{y}|\mathbf{z_0}=\mathbf{Z_0},\mathbf{z_1}=\mathbf{Z_1})$, then a PixMix blending stage can be described by sampling from the mixture distribution $q(\mathbf{y}|\mathbf{z_0},\mathbf{z_1})=\EX_k\left[g_k(\mathbf{y}|\mathbf{z_0},\mathbf{z_1})\right]$.  Specifically, PixMix considered two blending methods: an arithmetic mean and a geometric mean.  In addition to the mixture generalization, PixMix diverged from MixUp by considering conic weight combinations to further increase diversity.  Although the sum of the weights used in PixMix was not strictly equal to 1, the sampling used ensured that the expected value of the sum would still remain 1.  PixMix also introduced fractal images into their augmentation pipeline.  Since fractals were unlikely to collide with the data manifold in a label-conflicting manner, they could be used for blending to achieve a large increase in diversity with minimal downsides.

IPMix \cite{huangIPMixLabelpreservingData2024} built on PixMix \cite{hendrycksPixMixDreamlikePictures2022} through considering a variety of pipeline structures, and increasing the diversity of blending stages further by considering more blending methods.  In addition to the arithmetic and geometric means considered by PixMix, IPMix also used pixel level and joint pixel-channel (element) mixing.  With these adjustments, IPMix enhanced sample diversity and improved model performance. These performance increases were particularly notable on robustness benchmarks.

\subsection{Robust Deep Learning} %
Ensuring the safety of \acrshort{dl} systems is a critical aspect of deploying models in real-world applications, particularly in high-stakes environments. The risks associated with unsafe ML deployment, as highlighted in prior research~\cite{Finlayson2018AdversarialAA, Papernot2018SoKSA, Skaf2020ApplyingNA}, include severe economic, societal, and ethical consequences. With the advent of Self-Driving Cars~\cite{hwangEMMAEndtoEndMultimodal2024, pengImprovingAgentBehaviors2025} and \acrfull{lmm}~\cite{Huang2023LanguageIN, Shen2023HuggingGPTSA, OpenAI:2023ktj, touvronLLaMAOpenEfficient2023}, safety concerns have taken center stage, as these models, despite their impressive capabilities are prone to errors and can confidently provide incorrect predictions or fail under adversarial questioning. To address such challenges, various safety measures have been proposed, encompassing robustness, calibration, and anomaly detection, among others. Building on Hendricks \etal~\cite{hendrycksPixMixDreamlikePictures2022}, we categorize safety metrics tasks into four key subdomains, discussed below.

\textbf{Robustness.}  
Robustness in \acrshort{ml} systems pertains to their ability to maintain performance under distributional shifts or adverse conditions. Corruption robustness, for instance, evaluates resistance to natural perturbations encountered in real-world settings. ImageNet-C~\cite{hendrycksBenchmarkingNeuralNetwork2018} benchmark, a variant of ImageNet, introduces 15 common corruptions across five levels of severity, serving as a benchmark for assessing models' robustness under challenging, real-world conditions~\cite{Hendrycks2020TheMF_imagenet_r}
It is commonly used as a difficult, held-out test set for models trained on the ImageNet dataset~\cite{dengImageNetLargeScaleHierarchical2009}. 
Further extending the robustness benchmarks, Mintun \etal proposed ImageNet-$\overline{\text{C}}$~\cite{Mintun2021OnIB}, a complementary set of corruptions for evaluation. 
ImageNet-R benchmark~\cite{Hendrycks2020TheMF_imagenet_r} evaluates abstract visual generalization capability beyond natural corruptions to test model performance against diverse object renditions, such as paintings, cartoons, graffiti, embroidery, origami, sculptures, and toys, reflecting broader variations encountered in real-world scenarios. The dataset contains 200 categories instead of the original 1000 total categories in the ImageNet dataset.
Hendrycks \etal also introduced ImageNet-P \cite{hendrycksBenchmarkingNeuralNetwork2018}, which measures prediction consistency under non-adversarial input perturbations on 10 different perturbation types.

\textbf{Adversarial robustness.} It addresses the challenge of imperceptible image perturbations crafted to mislead models~\cite{Dong2017BoostingAA}. 
Studies have noted a trade-off between robustness to adversarial perturbations and accuracy on clean data~\cite{Xie2018FeatureDF, Xie2019AdversarialEI}. 
In the domain adaptation context, Bashkirova \etal~\cite{bashkirova2021visda} explored robustness during test-time adaptation and anomaly detection~\cite{ruff2021unifying}. 
Additionally, Yin \etal~\cite{Yin2019AFourierPerspective} observed that adversarial training, while improving performance under adversarial settings, can degrade robustness against certain corruptions. They suggest that this vulnerability is partly due to the model's reliance on spurious correlations~\cite{Sagawa2020Distributionally, koh2021Wilds}.
Geirhos \etal~\cite{Geirhos2018ImageNettrainedCA} highlight the texture bias present in \acrshort{cnn} and demonstrate that training with a diverse set of stylized images can enhance robustness against these shifts. 
Ensuring robustness in \acrshort{ml} models involves making them resilient to various forms of data shifts that may occur during testing.

\textbf{Calibration.}  
Calibration refers to the alignment between a model’s predicted confidence and its actual accuracy. Properly calibrated predictions are vital in real-world applications where overconfidence can lead to critical errors. Bayesian approaches~\cite{Guo2017OnCO} are widely employed for estimating uncertainty and improving calibration. Recalibration methods, such as those proposed by Kuleshov \etal~\cite{Kuleshov2018AccurateUF}, address the miscalibration of credible intervals to ensure reliability. Ovadia \etal~\cite{Ovadia2019CanYT} provide a comprehensive evaluation framework for assessing models' calibration and accuracy under distributional shifts, highlighting the importance of confidence estimation in dynamic environments.

These advancements highlight the versatility and importance of data augmentation across a wide range of deep learning applications, underscoring its central role in enhancing model performance, particularly in challenging scenarios involving data corruption or domain shifts. By addressing robustness and calibration aspects, ML safety frameworks are better equipped to mitigate risks, ensuring the deployment of reliable and trustworthy models in real-world scenarios.

\section{Methodology}\label{sec:methodology}
We introduce LayerMix, a layering-based data augmentation framework designed to balance affinity and diversity and increase both clean accuracy and safety metrics.  In our design, we find several areas in previous work where diversity can be eliminated without significantly impacting model performance and reallocate this in new ways to achieve better results.  In particular, we introduce covariance between augmentation stages in our pipeline, we employ augmentations on fractals for mixing, we redesign the augmentation pipeline, and reweigh the blending methods used across PixMix \cite{hendrycksPixMixDreamlikePictures2022} and IPMix \cite{huangIPMixLabelpreservingData2024}.

\subsection{Pipeline Covariance}
\label{sec:meth.cov}
We determined a simple method for introducing covariance into the augmentation pipeline.  We start by sampling a single transformation $f_k$ from a set of possible transformations.  Next, we apply the selected transformation $f_k$ for each augmentation stage in our pipeline.  Mathematically, this is equivalent to first sampling $k$ from a multinomial distribution, then sampling $\mathbf{Y}\sim f_k(\mathbf{y}|\mathbf{z})$ to apply each augmentation stage in the pipeline.  To analyze the covariance between stages, we construct the joint distribution over all stages' input-output spaces $\mathbf{x_n}$, where $n$ is the index of the stage in the pipeline.  Unlike the \acrshort{iid} approach described in \cref{sec:background.augpipe}, stages are only independent after conditioning on the outcome $k$ of the multinomial distribution, allowing for the direct construction of the conditional distribution $p_{layermix}(\mathbf{x}|k) = \prod_nf_k(\mathbf{x_n})$.  Taking the expected value over $k$ yields the marginal distribution over $\mathbf{x}$:

\begin{equation}
    p_{layermix}(\mathbf{x}) = \EX_{k}\left[\prod_nf_k(\mathbf{x_n})\right]
    \label{eqn:layer-joint-dist}
\end{equation}
\Cref{eqn:layer-joint-dist} is notably different from the corresponding marginal distribution in \acrshort{iid} pipelines.  Since each stage is marginally independent $p_{iid}(\mathbf{x_n}, \mathbf{x_m}) = p_{iid}(\mathbf{x_n})p_{iid}(\mathbf{x_m}) \forall n\neq m$, the joint distribution $p_{iid}$ can be expressed as the product over the per-stage mixture distributions:

\begin{equation}
    p_{iid}(\mathbf{x}) = \prod_n\EX_{k}\left[f_k(\mathbf{x_n})\right]
    \label{eqn:iid-joint-dist}
\end{equation}

Using \cref{eqn:layer-joint-dist}, the auto-covariance matrix over $\mathbf{x}$ can be calculated analytically.  Throughout this calculation we consider $x_n\in\mathbb{R}$, the general case follows.  To perform this calculation, we first consider the diagonal entries, $\mathbf{K_{X_iX_{i}}}$, where $p(\mathbf{x})=p_{layermix}(\mathbf{x}) $ for brevity:

\begin{equation}
    \begin{split}
        \mathbf{K_{X_iX_{i}}}
            &= \EX\left[X_i^2\right] - \EX\left[X_i\right]^2 \\
            &= \left\langle x_i^2,p(x_i)\right\rangle - \left\langle x_i,p(x_i)\right\rangle^2 \\
            &= \EX_k\left[\left\langle x_i^2,f_k(x_i)\right\rangle\right] - \EX_k\left[\left\langle x_i,f_k(x_i)\right\rangle\right]^2 \\
            &= \EX_k\left[\mu_{ki}^2 + \sigma_{ki}^2\right] - \EX_k\left[\mu_{ki}\right]^2 \\
    \end{split}
    \label{eqn:layer-autocov-diag}
\end{equation}

As a second case, we calculate the off-diagonal entries, $\mathbf{K_{X_iX_{j\neq i}}}$, where $p(\mathbf{x})=p_{layermix}(\mathbf{x}) $ for brevity:

\begin{equation}
    \begin{split}
        \mathbf{K_{X_iX_{j\neq i}}}
            &= \EX\left[X_iX_j\right] - \EX\left[X_i\right]\EX\left[X_j\right] \\
            &= \left\langle x_ix_j, p(x_i,x_j)\right\rangle - \left\langle x_i,p(x_i)\right\rangle\left\langle x_j,p(x_j)\right\rangle \\
            &= \EX_k\left[\left\langle x_ix_j,f_k(x_i)f_k(x_j)\right\rangle\right] - \EX_k\left[\left\langle x_i,f_k(x_i)\right\rangle\right]\EX_k\left[\left\langle x_j,f_k(x_j)\right\rangle\right] \\
            &= \EX_k\left[\left\langle x_i,f_k(x_i)\right\rangle\left\langle x_j,f_k(x_j)\right\rangle\right] - \EX_k\left[\left\langle x_i,f_k(x_i)\right\rangle\right]\EX_k\left[\left\langle x_j,f_k(x_j)\right\rangle\right] \\
            &= \EX_k\left[\mu_{ki}\mu_{kj}\right] - \EX_k\left[\mu_{ki}\right]\EX_k\left[\mu_{kj}\right] \\
    \end{split}
    \label{eqn:layer-autocov-offdiag}
\end{equation}

Using \cref{eqn:layer-autocov-diag} for the diagonal elements of $\mathbf{K_{X_iX_j}}$ and \cref{eqn:layer-autocov-offdiag} for the off-diagonal elements results in the following complete auto-covariance matrix expression:

\begin{equation}
    \mathbf{K_{X_iX_j}} =
    \begin{cases*}
        \EX_k\left[\sigma_{ki}^2\right] + \EX_k\left[\mu_{ki}^2\right] - \EX_k\left[\mu_{ki}\right]^2 & if i = j \\
        \EX_k\left[\mu_{ki}\mu_{kj}\right] - \EX_k\left[\mu_{ki}\right]\EX_k\left[\mu_{kj}\right] & else
    \end{cases*}
    \label{eqn:layer-autocov}
\end{equation}

The result of this calculation is a relation between the covariance of any two joint input-output image spaces $\text{Cov}(X_i,X_j)$ and low-order statistics of the transformations $f_k$.  This covariance structure can be directly contrasted against the implied covariance structure from the \acrshort{iid} pipelines defined by \cref{eqn:iid-joint-dist}: $\mathbf{K_{X_iX_j}} = \text{diag}(\EX_k[\sigma_{k}^2])$.
This approach decreases sample diversity without hindering performance.  In combination with this, we increase the magnitude of the transformations $f_k$ to achieve an overall increase in model performance.

\subsection{Fractals Augmentation}
\label{sec:meth.fractals}
Fractals can be created through various methods, with iterated function systems being one of the most widely used techniques~\cite{Anderson2021ImprovingFP}. However, it demands a significant research and development effort. PixMix~\cite{hendrycksPixMixDreamlikePictures2022} simplified it by collecting a total of $14,\!230$ colored fractals from curated repositories on DeviantArt\footnote{\url{https://www.deviantart.com/}} instead of generating a custom set of diverse fractals. They used these fractals to increase the dataset diversity by simply resizing them, followed by a random crop to match the input image size. It was later used by IPMix~\cite{huangIPMixLabelpreservingData2024} as well. 

Kataoka \etal~\cite{kataokaPretrainingNaturalImages} noticed complexity of the fractals provided by shapes and contours is only needed for pertaining a \acrshort{cnn} as colors do not provide additional information~\cite{Anderson2021ImprovingFP}. To utilize the complexity of the fractals effectively and match with ~\cite{kataokaPretrainingNaturalImages}, we used a gray-scale version of the $14,\!230$ fractals (collected by PixMix \cite{hendrycksPixMixDreamlikePictures2022}) for blending. We enhanced the diversity of fractals by randomly flipping them horizontally and vertically during training. Additionally, we observed that utilizing gray-scale fractals with earlier methods led to a marked improvement in results (details are provided in \cref{subsubsec:cifar_experiments}). \Cref{fig:grayscale_fractals} shows samples of gray-scale fractals. 

\subsection{Reweighted Blending}
\label{sec:meth.blend}
As mentioned in \cref{sec:background.augpipe}, the blending stage of an augmentation pipeline can be mathematically expressed as sampling from a mixture distribution $q$ over blending methods $g_k$: $q(\mathbf{y}|\mathbf{z_0},\mathbf{z_1})=\EX_k\left[g_k(\mathbf{y}|\mathbf{z_0},\mathbf{z_1})\right]$, where $k$ indexes the blending method.
The distribution of blending methods is not specified analytically but rather implied from the definition of the random variable $\mathbf{Y}$ in terms of the input images $\mathbf{Z_0}$ and $\mathbf{Z_1}$, along with other random variables sampled from known distributions.  These expressions are shown in \cref{tab:blending_exprs}.

\begin{table}[h]
    \centering
    \caption{Blending expressions. $\odot$ represents the Hadamard product}
    \begin{tabular}{l|c|c}
        \toprule
        Blending Method & Expression & Probability \\
        \midrule
        Arithmetic Mean & $\mathbf{Y}=a\mathbf{Z_0}+b\mathbf{Z_1}$ & 33.3\% \\
        Geometric Mean  & $\mathbf{Y}=2^{a+b-1}\cdot\mathbf{Z_0}^a\mathbf{Z_1}^b$ & 33.3\% \\
        Pixel Mixing    & $\mathbf{Y}=\mathit{M}\odot\mathbf{Z_0} + (1-\mathit{M})\odot\mathbf{Z_1}$ & 16.6\% \\
        Element Mixing  & $\mathbf{Y}=\mathit{M}\odot\mathbf{Z_0} + (1-\mathit{M})\odot\mathbf{Z_1}$ & 16.6\%\\
        \bottomrule
    \end{tabular}
    \label{tab:blending_exprs}
\end{table}

For the arithmetic and geometric mean expressions (shown in \cref{tab:blending_exprs}), both $a$ and $b$ are random variables.  In both expressions, $a\sim \frac{1}{2}\mathbf{B}(x|\beta,1) + \frac{1}{2}\mathbf{B}(x-1|1,\beta)$, and $b\sim \frac{1}{2}\mathbf{B}(x|1,\beta) + \frac{1}{2}\mathbf{B}(-x|1,\beta)$, where $\mathbf{B}$ represents the Beta-Distribution.  This sampling results in the conic combination proposed by PixMix \cite{hendrycksPixMixDreamlikePictures2022} and provides a parameter $\beta$, the blending ratio, which can be tuned to adjust these blending methods.  In the case of pixel and element mixing, $\mathit{M}$ is a boolean mask that determines whether a pixel value is taken from $\mathbf{Z_0}$ or $\mathbf{Z_1}$ as defined by IPMix \cite{huangIPMixLabelpreservingData2024}.

Through extensive ablation studies, we observed that the arithmetic and geometric blending methods are particularly effective in enhancing clean accuracy by promoting the learning of diverse and discriminative features.
Conversely, pixel-wise mixing significantly improves robustness to corruption, likely due to its localized blending effect.
To balance these strengths, we adjust the multinomial weights used to construct the mixture distribution $q$, favoring the arithmetic and geometric methods over the pixel and element methods.  In practice, we selected a blending method using the associated probabilities shown in \cref{tab:blending_exprs}. This weighting ensured a favorable bias towards methods that contributed to clean accuracy while retaining the robustness benefits of pixel-wise mixing.

\subsection{Augmentation Pipeline}
\label{sec:meth.augpipe}
\begin{figure}
    \centering
    \includegraphics[width=0.75\textwidth]{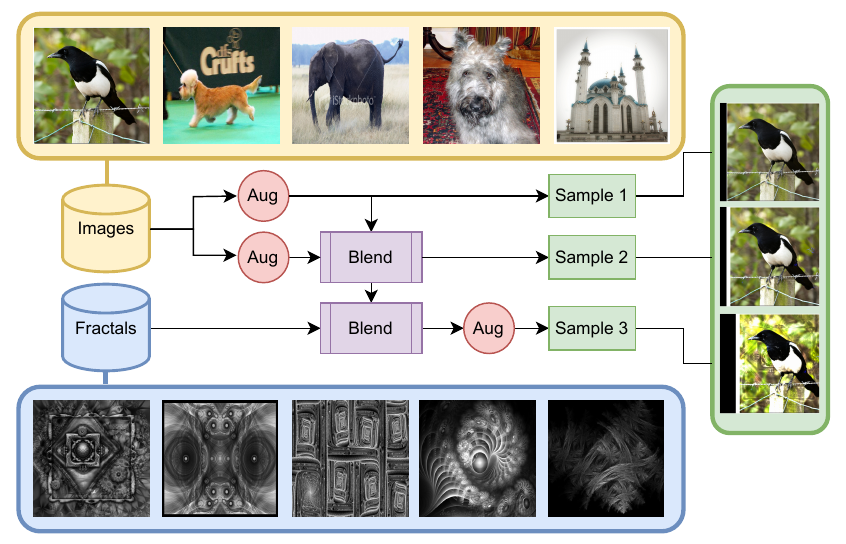}
    \caption{Complete Pipeline of LayerMix.  The resulting image produced by the LayerMix pipeline is uniformly selected from samples 1, 2, and 3 produced by the pipeline.  All \textbf{Aug} blocks are correlated by the covariance structure described in \cref{sec:meth.cov}.  All \textbf{Blend} blocks are independent and sample from the re-weighted blending mixture distribution described in \cref{sec:meth.blend}.}
    \label{fig:blockdiagram}
\end{figure}

The LayerMix pipeline uses a combination of correlated augmentation stages and blending stages.  A diagram of the pipeline is shown in \cref{fig:blockdiagram}, and the pseudo-code is shown in Code Block \ref{code:pseudo_code}.  The final sample produced by LayerMix is selected uniformly between samples 1, 2, and 3 in \cref{fig:blockdiagram}.  We use samples from each of these layers in our pipeline to enable control over the distribution of the diversity of samples produced by our pipeline.  Intuitively, sample 1 will have less diversity than sample 2, which in turn will have less diversity than sample 3.  By randomly selecting between each of these samples, we are able to reduce the average deviation from the data manifold while still producing highly diverse samples.

\begin{center}
\begin{minipage}{0.78\linewidth}
\begin{lstlisting}[style=python,caption=LayerMix Pipeline., label=code:pseudo_code]
def layermix(img, mixing_pic) -> Tensor:
    step = random.randint(3)
    img_copy = img.clone()
    aug_fn = random.choice(aug_fns)
    img = aug_fn(img, magnitude)
    if step == 0: return img

    img_2 = aug_fn(img_copy, magnitude)
    blending = random.choice(blending_fns)
    img = torch.clip(blending(img, img_2, blending_ratio), 0, 1)
    if step == 1: return img
    
    blending = random.choice(blending_fns)
    img = torch.clip(blending(img, mixing_pic, blending_ratio), 0, 1)
    return aug_fn(img, magnitude)
\end{lstlisting}
\end{minipage}
\end{center}

The fractals in \cref{fig:blockdiagram} are sampled uniformly and augmented using the strategy defined in \cref{sec:meth.fractals}.
Through a combination of the intrinsic diversity present in fractals alongside the augmentations, the third sample in the LayerMix pipeline will produce notably more diverse samples than the first two.
We find that this combination works well during training as sample 3 is able to effectively blend out the edges of the data manifold over which the model is learning.
To enable further control over the diversity of the samples produced by the LayerMix pipeline, we follow \cite{cubukRandAugmentPracticalAutomated2019} and parameterize most of our transformations $f_k$ by an augmentation magnitude parameter $m$.  The transformations are tuned such that any value of $m$ results in a similar increase in sample diversity across all transformations $f_k$.
This tuning was performed by \cite{cubukRandAugmentPracticalAutomated2019}
and results in the per-transform value ranges shown in \cref{tab:image_augmentations}.

\input{tables/augs}

In the construction of our pipeline, we specifically avoided applying two augmentation stages sequentially.  This is due to a form of linearity that emerges from the covariance structure (\cref{sec:meth.cov}) when combined with the implementation of the transformations $f_k$.  Mathematically, for many values $k$, $f_k(\mathbf{y}|\mathbf{z}=\mathbf{Z}\sim f_k(\mathbf{y}|\mathbf{z};am);bm) \approx f_k(\mathbf{y}|\mathbf{z};(a+b)m)$.  Intuitively, if $f_k$ represents a rotation by $a\sim\mathcal{N}(0,m)$ degrees, then the result of applying $f_k$ twice would be a rotation by $a\sim\mathcal{N}(0,2m)$ degrees.  While this is not a precise description since not all transformations have parameters that can be linearly combined and parameters are not always sampled according to a normal distribution, it can be extended to show the approximate linearity for most of the transformations described in \cref{tab:image_augmentations}.  Hence, we utilize blending stages as a form of non-linearity to avoid augmentation stages from collapsing in our pipeline.

We have extended these ideas by exploring new strategies for label-preserving augmentations, further improving model generalization and corruption resistance. Our proposed approach draws inspiration from PixMix~\cite{hendrycksPixMixDreamlikePictures2022} and IPMix~\cite{huangIPMixLabelpreservingData2024} but introduces novel techniques to leverage layered and effective information fusion strategies. By integrating diverse patterns from multiple layers, our method enables the training of models that are more robust to data corruption and domain shifts, pushing the boundaries of existing augmentation techniques. \Cref{fig:layermix_comp} shows visual differences in different augmentation pipelines. Fractals and visual samples generated by LayerMix are presented in \cref{sec:more_samples}.

\section{Experiments}\label{sec:experiments}
To assess the performance of our proposed method LayerMix, we have used a similar experimentation setting as PixMix~\cite{hendrycksPixMixDreamlikePictures2022} on benchmark datasets i.e., CIFAR-10, CIFAR-100~\cite{krizhevskyLearningMultipleLayers2009}, and ImageNet-1K~\cite{russakovskyImagenetLargeScale2015} for training and evaluation of the models on classification accuracy. We have also done rigorous experiments on ImageNet-200~\cite{Hendrycks2020TheMF_imagenet_r} dataset. We also evaluated the robustness of performance across various safety-related tasks on corrupted versions of the test sets. For a fair comparison to earlier methods, any difference to the experimentation setting of PixMix is especially highlighted. IPMix~\cite{huangIPMixLabelpreservingData2024} uses \acrshort{jsd} Loss proposed by Augmix~\cite{hendrycks*AugMixSimpleData2019} instead of standard \acrfull{bce} Loss \cite{goodTerminologyNotationInformation1956}. It uses 3 times the memory but provides better convergence. Therefore, we trained IPMix without \acrshort{jsd} Loss for a fair comparison. We have highlighted the performance improvement with \acrshort{jsd} loss for LayerMix as well.

We benchmark our approach against state-of-the-art data augmentation techniques, such as MixUp~\cite{Zhang2018mixupBE}, CutMix~\cite{Yun2019CutMixRS}, AugMix\cite{hendrycks*AugMixSimpleData2019}, RandomErase~\cite{Zhong2017RandomED} and recent methods, inclunding PixMix \cite{hendrycksPixMixDreamlikePictures2022} and IPMix\cite{huangIPMixLabelpreservingData2024}. Standard augmentations such as random cropping and flipping serve as baseline. These methods vary in complexity, with strategies ranging from pixel-wise interpolations (MixUp) to advanced augmentation pipelines (AugMix, IPMix). PixMix leverages neural style transfer and augmentation sampling to enhance robustness, requiring fewer augmentations per image than AugMix. IPMix introduces a hierarchical blending strategy inspired by iterative and progressive mixing, offering improvements in both clean accuracy and robustness metrics.

\subsection{Evaluation Benchmarks and Metrics}
We evaluate our methodology using the benchmark datasets CIFAR-10, CIFAR-100~\cite{krizhevskyLearningMultipleLayers2009}, ImageNet-200~\cite{Hendrycks2020TheMF_imagenet_r}, and ImageNet-1K~\cite{russakovskyImagenetLargeScale2015} dataset, alongside their respective robustness benchmarks across 4 distinct \acrshort{ml} safety tasks. All methods are initially trained on the clean versions of CIFAR-10, CIFAR-100, and ImageNet datasets before being tested on the tasks outlined below.

\subsubsection{Corruption Robustness:} In this task, the objective is to classify corrupted images from the CIFAR-10-C, CIFAR-100-C, and ImageNet-C datasets (introduced by \cite{hendrycksBenchmarkingNeuralNetwork2018}). To quantify performance, we use the mean corruption error (mCE) metric, which measures the classification error averaged across all 15 (+4 supplementry) corruption types and 5 levels of severity for each type. A lower mCE indicates better robustness to corruption. We also evaluated on the CIFAR-10-$\overbar{\text{C}}$, CIFAR-100-$\overbar{\text{C}}$, ImageNet-200-$\overbar{\text{C}}$, ImageNet-1k-$\overbar{\text{C}}$ datasets, introduced by~\cite{Mintun2021OnIB}, that has 10 additional corruptions for robustness evaluation.

Furthermore, supplementary datasets such as ImageNet-R~\cite{Hendrycks2020TheMF_imagenet_r} dataset measure robustness to rendition variations such as art, cartoons,  graffiti, embroidery, toys, and video game renditions of ImageNet classes. ImageNet-R has renditions of 200 ImageNet classes, resulting in 30,000 images.
The datasets span diverse settings, enabling a comprehensive evaluation of clean accuracy, robustness to corruptions, prediction consistency, calibration, and anomaly detection.

\subsubsection{Prediction Consistency:} This task aims to ensure the consistent classification of sequences of perturbed images from the CIFAR-100-P and ImageNet-P datasets where each sample is a sequence of images undergoing gradual perturbations such as zoom, translation, or brightness changes. These sequences enable assessing stability in predictions under minor shifts. The primary metric used here is the mean flip probability (mFP) introduced by \cite{hendrycksBenchmarkingNeuralNetwork2018}, which represents the likelihood that adjacent frames in a sequence are assigned different predicted classes. Mathematically, this can be expressed as:
\begin{equation}
    \text{mFP} = \mathbb{P}_{x \sim S} \left(f(x_i) \neq f(x_{i-1})\right),
\end{equation}
where $x_i$ denotes the $i$-th image in a temporal sequence. For non-temporal sequences, such as those with progressively increasing noise levels within sequence $S$, the metric is adapted as:
\begin{equation}
    \text{mFP} = \mathbb{P}_{x \sim S} \left(f(x_i) \neq f(x_1)\right).
\end{equation}
In both cases, a lower mFP indicates greater consistency in the model's predictions. Sometimes it is also aided by the consistency of top-5 predictions under perturbations. We define a distance metric \( d(\tau(x), \tau(x')) \) between the permutations \( \tau(x) \) and \( \tau(x') \) representing the ranked predictions of a model \( f \). This metric penalizes changes in the top-5 predictions as follows:
\begin{equation}
d(\tau(x), \tau(x')) = \sum_{i=1}^5 \sum_{j=\min\{i,\sigma(i)\}+1}^{\max\{i,\sigma(i)\}} \mathds{1}(1 \leq j - 1 \leq 5),
\end{equation}
\noindent
where \( \sigma = (\tau(x))^{-1}\tau(x') \). The mean Top-5 Distance (mT5D) is the average of \( \text{T5D}^f_p \) values across all perturbation sequences, providing a measure of top-5 consistency under perturbations.

\subsubsection{Adversarial Robustness:} In this task, the objective is to classify images that have been adversarially perturbed using the \acrfull{pgd} method \cite{madry2018towards}. Specifically, we consider untargeted attacks on CIFAR-10 and CIFAR-100, with an $\ell_\infty$ perturbation budget of $2/255$ and 20 optimization steps. 
The performance metric for this task is the classifier error rate, where a lower error rate indicates better robustness to adversarial perturbations. We do not include results for ImageNet models in our tables, as all tested methods experience a complete drop in accuracy to zero under this attack budget. This was also observed by PixMix\cite{hendrycksPixMixDreamlikePictures2022}. 

\subsubsection{Calibration:} This task evaluates the model’s ability to produce calibrated prediction probabilities that align with the empirical frequency of correctness. For instance, if a model predicts a 70\% probability of rain on ten occasions, we expect it to be correct approximately 7 out of those 10 times. Formally, we aim for the model’s posterior probabilities to satisfy:
\begin{equation}
    \mathbb{P} \left(Y = \arg \max_i f(X)_i \mid \max_i f(X)_i = C\right) = C
\end{equation}
where $X$ and $Y$ are random variables representing the data distribution, and $f(X)_i$ denotes the predicted probability for class $i$. To quantify calibration, we use the root mean square (RMS) calibration error \cite{hendrycks2018deep}, defined as:
\begin{equation}
    \text{RMS Calibration Error} = \sqrt{\mathbb{E}_C \left[\left(\mathbb{P}(Y = \hat{Y} \mid C = c) - c\right)^2\right]}
\end{equation}
where $C$ represents the model’s confidence that its predicted label $\hat{Y}$ is correct. We employ adaptive binning \cite{nguyen15calibration} to compute this metric, with lower values indicating better calibration. We calculate the calibration using all the robustness datasets.

\subsection{CIFAR Results}
\subsubsection{Training Setup.}
For the CIFAR experiments, we employ a WideResNet-40-4 (WRN-40-4) architecture \cite{zagoruykoWideResidualNetworks2016} with a dropout rate of 0.3. The initial learning rate is set to 0.1 and is adjusted throughout training using a cosine annealing schedule \cite{loshchilov2017sgdrCosineAnnealing} with a weight decay of $0.0005$. For the LayerMix experiments, we set the hyperparameters $magnitude=8$ and $blending\_ratio=3$. This setup ensures a fair comparison across the different augmentation strategies. We used $\beta=3$ and $k=3$ for all PixMix and IPmix experiments for CIFAR-10/100. All CIFAR experiments were conducted using one RTX3060 \gls{gpu}. 

\subsubsection{Experiments.}\label{subsubsec:cifar_experiments}
\begin{figure}
    \centering
    \includegraphics[width=1\linewidth]{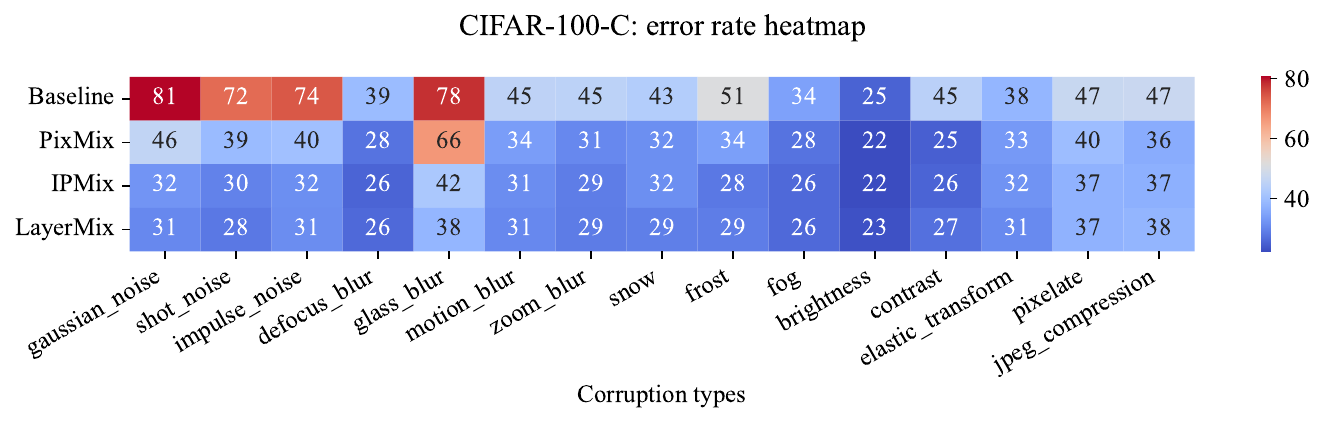}
    \caption{Corruption Error values for various methods on CIFAR-100-C.}
    \label{fig:cifar100_c_heatmap}
\end{figure}

\input{tables/cifar/horizontal_combine_results_cifar10_100}
As shown in \cref{tab:cifar_10_100_horizontal}, our proposed method, LayerMix, consistently outperforms the standard baseline as well as other state-of-the-art methods across all evaluated safety metrics. LayerMix performs better than all other approaches on tasks such as Corruption, Consistency, and Calibration. Notably, it significantly enhances confidence calibration, achieving exceptionally low calibration errors on CIFAR-100. Regarding corruption robustness, the improvements on CIFAR-100-C and CIFAR-100-$\overbar{\text{C}}$ are particularly substantial, with the mean corruption error (mCE) reduced by 5.2\% and 3.3\% respectively relative to PixMix and by 19.7\% and 14.1\% compared to the baseline. Prediction consistency, measured via mFP, shows that our method achieves the lowest flip rates compared to PixMix and IPMix. Calibration performance is significantly enhanced. Our Method improves Adversarial robustness experiments against baseline but is second best overall against \acrshort{pgd} attacks in CIFAR-10. Detailed results on all the metrics on CIFAR-100 are presented in \cref{tab:full_results_cifar_100}. We also analyzed individual \acrfull{ce} values for various methods on CIFAR-100-C in \cref{fig:cifar100_c_heatmap} and noticed LayerMix improves on all the corruption types.
\input{tables/cifar/full_results_cifar100}
\subsubsection{\acrshort{jsd} loss and Optimal Hyperparameters.}
We also compared our method with the experimentation setting of IPMix, which uses \acrshort{jsd} loss instead of \acrshort{bce} loss. We also compared PixMix and IPMix by using grayscale and colored fractals. As mentioned in the \cref{tab:jsd_vs_fractals}, just by using gray-scale fractals instead of colored fractals with PixMix, we can improve corruption robustness performance; however, it negatively impacts the clean accuracy. Gray-scale fractals improve the IPMix performance on classification, robustness and consistency. We also note that gray-scale fractals slightly hurt adversarial performance in all methods. LayerMix trained with \acrshort{jsd} loss can surpass IPMix with 1.12\% on clean accuracy and 2.77\% on Adversarial Error. Few other methods \cite{Kim2020PuzzleME, Kim2021CoMixupSG} trained models for 400, 600, or 1,200 epochs due to high regularization provided by augmentation methods, so we also trained LayerMix for 400 epochs and noticed significant performance improvement over all other methods. We also evaluated the robustness of our pipeline hyperparameters in different combinations (as shown in \cref{tab:mag_vs_blending}) and noticed different combinations of magnitude $m$ and blending ratio $\beta$ result in a slight improvement in metrics. However, the overall metrics remain in the same error margin of the \acrshort{cnn}.
\input{tables/cifar/jsd_vs_fractals}
\input{tables/cifar/mag_vs_br}

\subsubsection{Other Models.}
We have also trained WideResNet-28-10 (WRN-28-10) \cite{zagoruykoWideResidualNetworks2016}, ResNet-18 \cite{heDeepResidualLearning2016} and ResNext-28 \cite{Xie2017RESNEXT} models for comparison with different models. As suggested by the original authors of the respective models, WRN models were trained for 100 epochs, while ResNet-18 and ResNeXt-29 were trained for 200 epochs, keeping all other settings the same. Based on  \cref{tab:diff_models}, we note that ResNeXt-29, having 6.9 Million parameters trained on 200 epochs, surpasses the WRN-40-4 model with 8.9 Million parameters trained with 100 epochs, suggesting heavy augmentation pipelines require more epochs to generalize better. Furthermore, WRN-28-10, having more parameters, surpasses the WRN-40-4 model in similar settings. We compare PixMix and IPMix trained with colored fractals. At the same time, we use gray-scale fractals for LayerMix, surpassing all the other methods on robustness and consistency metrics while being comparable in other metrics.  
\input{tables/cifar/diff_models}

\subsection{ImageNet Results}
The ImageNet-1K dataset encompasses 1,000 categories derived from WordNet noun synsets, covering a broad range of objects, including fine-grained distinctions. It includes 1.28 million color images, commonly resized to a resolution of $224 \times 224$ pixels for our experiments. We use its validation set, containing 50,000 images, to measure clean accuracy. ImageNet-200 uses the same 200 classes as ImageNet-R \cite{Hendrycks2020TheMF_imagenet_r}, a subset of 200 classes of ImageNet. It is important to note that ImageNet-200 is not the same as Tiny-ImageNet-200 \cite{leTinyImagenetVisual2015}, which uses a different subset of classes.

\subsubsection{Training Setup.}
For the ImageNet experiments, we trained ResNet-50 architecture \cite{heDeepResidualLearning2016}. Following PixMix~\cite{hendrycksPixMixDreamlikePictures2022}, we also fine-tuned the pre-trained model for 90 epochs, as regularization methods require more epochs to converge. The initial learning rate is set to 0.01 and is adjusted throughout training using a cosine annealing schedule \cite{loshchilov2017sgdrCosineAnnealing} with a batch size of 256. For the LayerMix experiments, we set the hyperparameters $magnitude=8$ and $blending\_ratio=3$. We used $\beta=4$, $k=4$, and $magnitude=1$ for all PixMix and IPmix experiments. All ImageNet experiments were conducted using one RTX TITAN \gls{gpu}.

\subsubsection{Performance Analysis}
\Cref{tab:imagenet200} presents our comprehensive evaluation of ImageNet-200. Our findings demonstrate that LayerMix consistently surpasses conventional augmentation techniques across all safety metrics. This stands in notable contrast to alternative augmentation approaches, which occasionally underperform relative to baseline methods (standard cropping and flipping operations). Significantly, LayerMix represents the first augmentation strategy to achieve comprehensive Pareto improvements over baseline measurements across diverse safety criteria.

\begin{figure}
    \centering
    \includegraphics[width=1\linewidth]{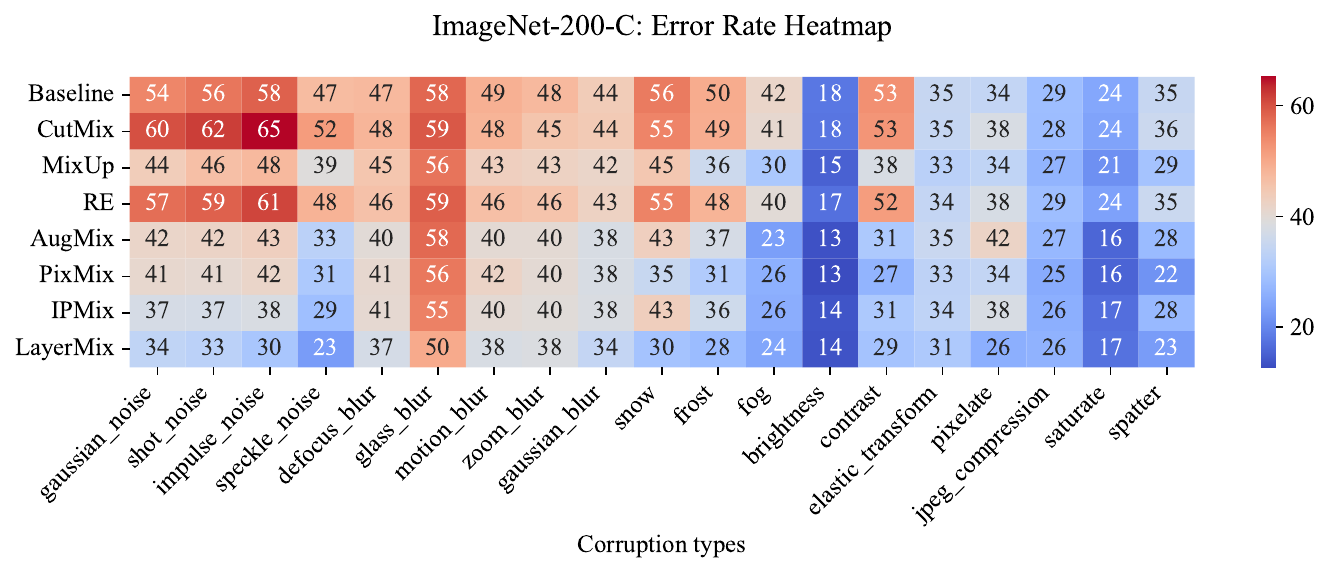}
    \caption{Individual Corruption Error (CE) values for various methods on ImageNet-200-C}
    \label{fig:imgnet_200_c_heatmap}
    \includegraphics[width=1\linewidth]{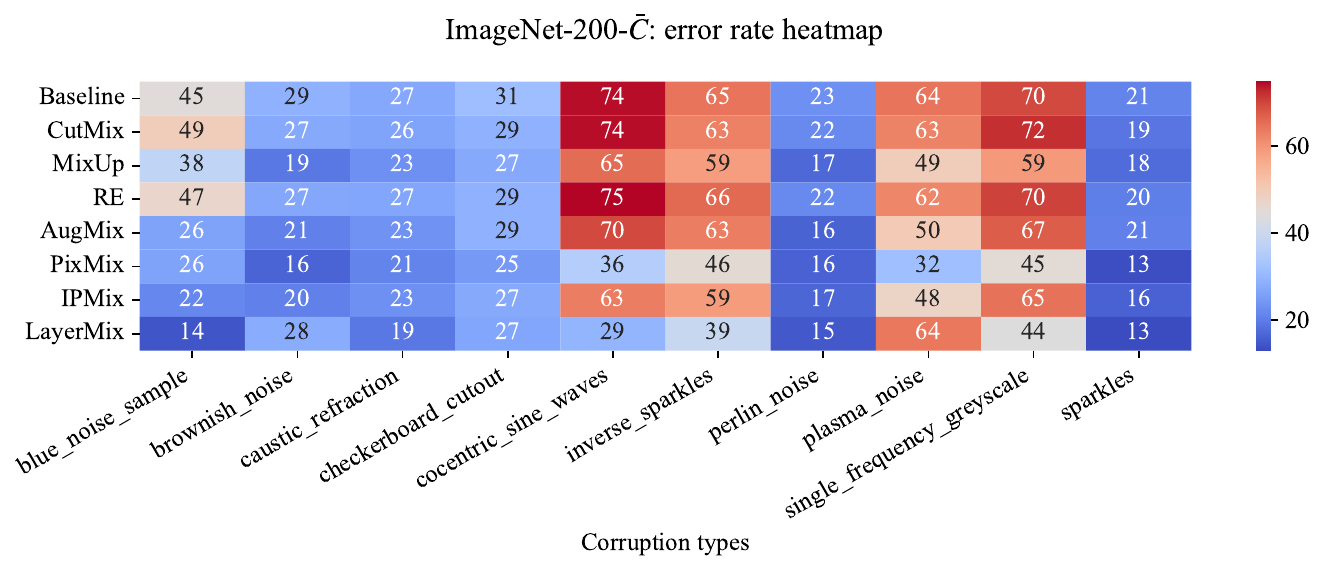}
    \caption{Corruption Error (CE) values for various methods on CIFAR-100-$\overbar{C}$.}
    \label{fig:imgnet_200_c_bar_heatmap}
\end{figure}

In terms of corruption robustness, LayerMix demonstrates exceptional performance, surpassing current state-of-the-art augmentation techniques. Specifically, we observed a 14.26\% improvement in mCE compared to the baseline and a 3.54\% enhancement over the next best method (PixMix). Regarding rendition robustness on ImageNet-R, LayerMix exhibited superior performance across all methods i.e., 8.13\% over baseline and 0.5\% over the next best. In contrast, LayerMix achieves a 0.5\% improvement in clean accuracy over baseline but has 0.24\% less than best(CutMix). LayerMix does outperform CutMix on all other metrics, highlighting its practical advantage for real-world applications where maintaining performance under ideal conditions remains crucial. We also provide the individual corruption improvement on Imagnet-200-C and ImageNet-200-$\overbar{\text{C}}$ in the form of heatmaps in \cref{fig:imgnet_200_c_heatmap} and \cref{fig:imgnet_200_c_bar_heatmap}. On ImageNet-1K, we observe similar trends as mentioned in \cref{tab:imagenet_1k}. 

\input{tables/imagenet/imagenet200}

Overall, the results highlight the efficacy of our approach in enhancing safety measures without compromising clean accuracy. These improvements make it an appealing choice for real-world applications where robustness and reliability are critical.

\input{tables/imagenet/full_results}

\section{Conclusion} \label{sec:conclusion}
This research introduces LayerMix, a novel and efficient data augmentation methodology that demonstrates significant advancement in \acrshort{ml} safety measures. The distinctive aspect of LayerMix lies in its innovative approach to augmentation complexity, specifically through the strategic integration of fractal patterns and feature visualizations into the training pipeline. LayerMix introduces covariance between augmentation stages of the pipeline.  In this work, we consider a particular structure selected due to its ease of implementation and improved results. This represents a departure from conventional augmentation techniques, offering a more sophisticated approach to model training. Our comprehensive evaluation framework encompassed multiple critical dimensions of \acrshort{ml} safety, including Corruption robustness, Rendition robustness, Prediction consistency, Adversarial robustness, and Confidence calibration. The empirical results demonstrate LayerMix's exceptional performance across this broad spectrum of safety metrics. 
These findings suggest promising directions for future research in robust \acrshort{ml} systems, particularly in applications where model reliability and safety are paramount. The success of LayerMix in simultaneously addressing multiple safety concerns while maintaining computational efficiency opens new avenues for developing increasingly robust and reliable \acrshort{ml} systems.
\subsection{Shortcomings and Future Directions}
While LayerMix demonstrates significant advancements, several limitations highlight opportunities for future research. 
One notable shortcoming is the availability of unique fractal patterns that generalize optimally across all datasets or domains. Additionally, the computational overhead introduced by the augmentation process, although efficient compared to similar methods, may still pose challenges to some. The scalability of LayerMix to large-scale datasets and models also warrants further investigation, particularly in scenarios where training efficiency is critical.

Future research could focus on extending the LayerMix framework to adaptively select or generate augmentation patterns during the training process. It could contrast different covariance structures between augmentation stages to determine the effects of our proposed structure.  Additionally, it is possible to introduce covariance between blending stages which was not investigated in this work.
The utilization of LayerMix in downstream applications, like object detection and segmentation needs to be investigated. Exploring the integration of LayerMix with unsupervised or self-supervised learning paradigms could further enhance its robustness and applicability. Moreover, addressing its computational demands through hardware-aware optimizations or lightweight implementations would improve its usability in real-world applications. 
Expanding the evaluation framework to include additional safety metrics and diverse benchmarks could provide a more comprehensive understanding of its performance. These directions could collectively contribute to the evolution of robust \acrshort{ml} systems that are both scalable and versatile.

\section*{Supplementary Materials} 
All of our code, training runs, and experimental meta-data to reproduce results are available at \url{https://github.com/ahmadmughees/layermix}.

\section*{Authors Contributions}
\textbf{Hafiz Mughees Ahmad}:
Conceptualization, Methodology, Software, Investigation, Validation, Data curation, Formal analysis, Visualization, Resources, Writing - Original draft preparation. 
\textbf{Dario Morle}:
Investigation, Validation, Formal Analysis, Software, Writing - Original draft preparation, Conceptualization. 
\textbf{Afshin Rahimi}: Funding acquisition, Project administration, Resources, Supervision, Writing - Review \& Editing.
\section*{Funding}
This study is supported by IFIVEO CANADA INC., Mitacs through IT16094, and the University of Windsor, Canada.

\section*{Institutional Review}
Not applicable.

\section*{Informed Consent}
Not applicable.

\section*{Data Availability}
The data used in this study is fully open-source and publicly available and links to each item are mentioned in \cref{tab:data_sources}.
\input{tables/data}

\section*{Conflict of Interest}
Author Dario Morle was employed by the company IFIVEO CANADA Inc. The authors declare that the research was conducted in the absence of any commercial or financial relationships that could be construed as a potential conflict of interest.

\section*{Acknowledgement}
We like to thank Dr. Zaigham Zaheer from MBZUAI for the helpful discussion during this project. 

\bibliographystyle{IEEEtran}
\bibliography{IEEEabrv,references}

\input{supplementry}

\begin{IEEEbiography}[{
\includegraphics[width=1in,height
=1.25in, clip, keepaspectratio]{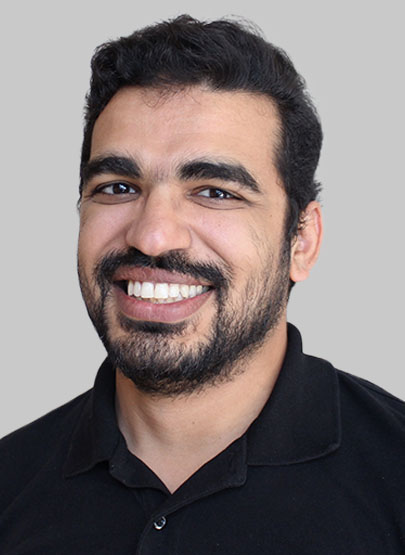}
}]
{Hafiz Mughees Ahmad} completed his Bachelor's and Master's in Electrical Engineering from the Institute of Space Technology, Pakistan, in 2015 and 2018, respectively. He is currently pursuing a Ph.D. at the University of Windsor, Canada. Alongside his studies, he serves as a Deep Learning Engineer at IFIVEO CANADA INC. His previous roles include Research Associate at Istanbul Medipol University, Turkey, and Lecturer at the Institute of Space Technology, Pakistan. His research focuses on Computer Vision and Deep Learning, with applications in Object Detection and real-time surveillance and monitoring in the production environment. He is a Graduate Student Member of IEEE.
\end{IEEEbiography}
\vskip -2\baselineskip plus -1fil
\begin{IEEEbiography}[{
\includegraphics[width=1in,height=1.25in, clip, keepaspectratio]{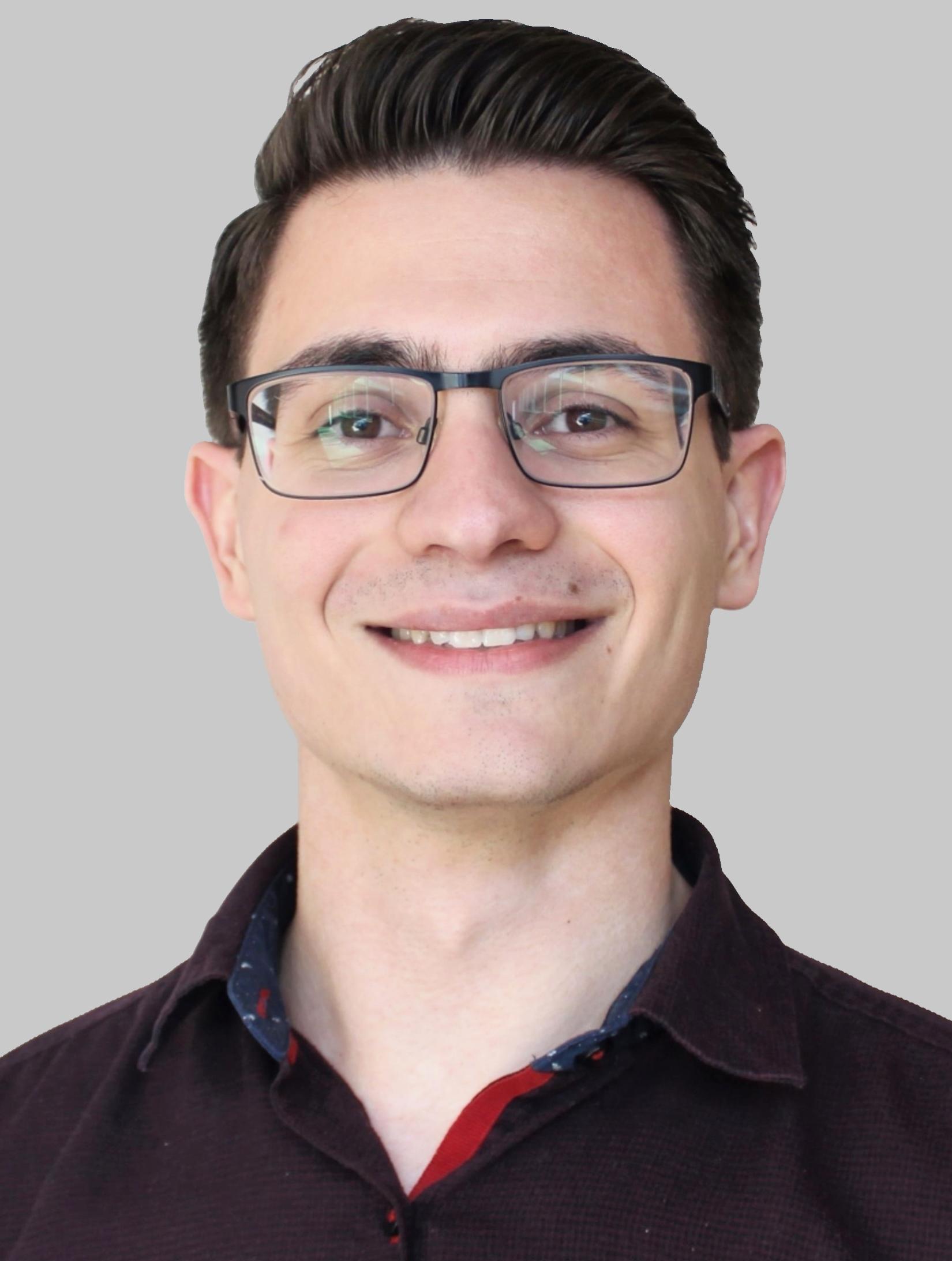}
}]
{Dario Morle} completed his Bachelor's in Electrical Engineering from the University of Windsor, Canada.  During his undergraduate degree, Dario assisted in research at both the Center for Computer Vision and Deep Learning, and the Human Systems Lab.  After graduating in 2022, he joined IFIVEO CANADA INC. as a Deep Learning Engineer where he worked on developing and deploying computer vision solutions for industrial applications.  His research interests include neural architecture design, energy-based models, and applications to industrial processes.
\end{IEEEbiography}
\vskip -2\baselineskip plus -1fil

\begin{IEEEbiography}
[{
\includegraphics[width=1in,height=1.25in,clip,keepaspectratio]{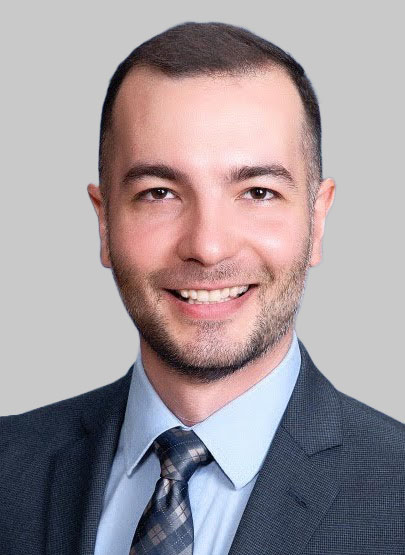}
}]
{Afshin Rahimi} received his B.Sc. degree from the K. N. Toosi University of Technology, Tehran, Iran, in 2010, and the M.Sc. and Ph.D. degrees from Toronto Metropolitan University, Toronto, ON, Canada, in 2012, and 2017, respectively, in Aerospace Engineering. He was with Pratt \& Whitney Canada from 2017 to 2018. Since 2018, he has been an Associate Professor in the Department of Mechanical, Automotive, and Materials Engineering at the University of Windsor, Windsor, ON, Canada. Since 2010, he has been involved in various industrial research, technology development, and systems engineering projects/contracts related to the control and diagnostics of satellites, UAVs, and commercial aircraft subsystems. In recent years, he has also been involved with industrial automation and using technologies to boost manual labor work in industrial settings. He is a senior member of IEEE, a lifetime member of AIAA, and a technical member of the PHM Society.
\end{IEEEbiography}
\vskip -2\baselineskip plus -1fil
\end{document}

%% file: tables/augs.tex
\definecolor{Added}{rgb}{1.0, 1.0, 0.13}
\begin{table}[ht]
\caption{List of transformations $f_k$.  Transformations with range constrained by the augmentation magnitude $m$.  We build upon TorchVision Transforms V2 \tablefootnote{\url{https://pytorch.org/vision/0.20/}} library for our experiments.}
\label{tab:image_augmentations}
\centering
\begin{tabular}{l|c||l|c||l|c }
    \toprule
    \textbf{Operation} & \textbf{Range} & \textbf{Operation} & \textbf{Range} & \textbf{Operation} & \textbf{Range} \\ \midrule
    equalize   &-& brightness & 0.1 $\rightarrow$ 1.9 & shear\_x& $-$0.3 $\rightarrow$ $+$0.3\\
    grayscale &-& posterize & 0 $\rightarrow$ 4 & shear\_y& $-$0.3 $\rightarrow$ $+$0.3 \\
    auto contrast   & -  &
    solarize   & 0 $\rightarrow$ 1 & translate\_x    & 0 $\rightarrow$ 0.33 (of image size) \\
    && rotate & $-30^\circ$ $\rightarrow$ $+30^\circ$ &translate\_y    & 0 $\rightarrow$ 0.33 (of image size)\\
    \bottomrule
\end{tabular}
\end{table}

%% file: tables/cifar/horizontal_combine_results_cifar10_100.tex
\begin{table*}[t]
\centering
\caption{CIFAR-10 and CIFAR-100 results with WRN-40-4 model for four distinct safety metrics. \textbf{Bold} is best, \underline{Underline} is second best.}
\label{tab:cifar_10_100_horizontal}
\begin{tabular}{l|r|ccccccccc} 
\toprule
  & & Baseline & Cutout & Mixup & CutMix & \makecell{Auto\\Augment} & AugMix & \textsc{PixMix} & IPMix & LayerMix\\
\midrule
\parbox[t]{1mm}{\multirow{4}{*}{\rotatebox{90}{CIFAR-10}}}
& Corruptions ($\downarrow$)& 26.4 & 25.9 & 21.0 & 26.5 & 22.2 & 12.4 & 12.1 & \underline{9.5} & \textbf{9.4} \\
& Consistency ($\downarrow$)& 3.4  & 3.7  & 2.9  & 3.5  & 3.6  & \underline{1.7}  & 1.9 & \textbf{1.6} & \textbf{1.6}\\
& Adversaries ($\downarrow$)& 91.3 & 96.0 & 93.3 & 92.1 & 95.1 & 86.8 & \textbf{83.2} & 86.8 & \underline{83.6}\\
& Calibration ($\downarrow$)& 22.7 & 17.8 & 12.1 & 18.6 & 14.8 & 9.4  & \underline{2.4} & 4.4 & \textbf{2.1} \\
\midrule
\parbox[t]{1mm}{\multirow{4}{*}{\rotatebox{90}{CIFAR-100}}}
& Corruptions ($\downarrow$)& 50.0 & 51.5 & 48.0 & 51.5 & 47.0 & 35.4 & 35.5 & \underline{30.8} & \textbf{30.3}  \\
& Consistency ($\downarrow$)& 10.7 & 11.9 & 9.5  & 12.0 & 11.2 & 6.5  &  6.3 & \underline{6.0}  & \textbf{5.6}   \\
& Adversaries ($\downarrow$)& 96.8 & 98.5 & 97.4 & 97.0 & 98.1 & 95.6 & \textbf{92.4} & \underline{95.0} & 95.5  \\
& Calibration ($\downarrow$)& 31.2 & 31.1 & 13.0 & 29.3 & 24.9 & 18.8 &  \underline{7.0} & 10.3 & \textbf{5.9}   \\
\bottomrule
\end{tabular}
\end{table*}

%% file: tables/cifar/full_results_cifar100.tex
\begin{table*}[ht]
\setlength\tabcolsep{5pt}
\small
\centering
\caption{Full results for CIFAR-100 with WRN-40-4 model. LayerMix with $m=8$ and $\beta=5$ does result in clean error of 2.4 but for standardization across all WRN-40-4 experiments, we report results of $m=8$ and $\beta=3$ here. \textbf{Bold} is best, \underline{Underline} is next best.} \label{tab:full_results_cifar_100}
\begin{tabular}{r | ccccccccc} 
\toprule
 & \multicolumn{1}{c}{Accuracy} & \multicolumn{2}{c}{Robustness} & \multicolumn{2}{c}{Consistency} & \multicolumn{1}{c}{Adversaries} & \multicolumn{3}{c}{Calibration} \\ 
 \cmidrule(lr){2-2} \cmidrule(lr){3-4} \cmidrule(lr){5-6} \cmidrule(lr){7-7} \cmidrule(lr){8-10}
 & Clean & C & $\overline{\text{C}}$ & \multicolumn{2}{c}{CIFAR-P} & PGD & Clean & C & $\overline{\text{C}}$ \\
 & \textcolor{gray}{Error} & \textcolor{gray}{mCE} & \textcolor{gray}{mCE} & \textcolor{gray}{mFP} & \textcolor{gray}{mT5D} & \textcolor{gray}{Error} & \textcolor{gray}{RMS} & \textcolor{gray}{RMS} & \textcolor{gray}{RMS} \\
\midrule
Baseline    & 21.3&	50.0&	52.0&	10.7&	2.7&	96.8&	14.6&	31.2&  30.9\\
Cutout      & 19.9&	51.5&	50.2&	11.9&	2.7&	98.5&	11.4&	31.1&  29.4\\
Mixup       & 21.1&	48.0&	49.8&	9.5&	3.0&	97.4&	10.5&	13.0&  12.9\\
CutMix      & \underline{20.3}&	51.5&	49.6&	12.0&	3.0&	97.0&	12.2&	29.3&  26.5\\
AutoAugment & \textbf{19.6}&	47.0&	46.8&	11.2&	2.6&	98.1&	9.9&	24.9&  22.8\\
AugMix      & 20.6&	35.4&	41.2&	6.5&	1.9&	95.6&	12.5&	18.8&  22.5\\
OE          & 21.9&	50.3&	52.1&	11.3&	3.0&	97.0&	12.0&	13.8&  13.9\\
PixMix      & 20.4& 35.5&   41.2&   6.3&    \underline{1.7}&    \textbf{92.4}&    \underline{7.0}&   \underline{10.5}& \textbf{11.3} \\
IPMix       & 20.4& \underline{30.8}&   \underline{39.4}&   \underline{6.0}&    \underline{1.7}&    \underline{95.0}&   10.3&   13.0& 17.8 \\
LayerMix    & 20.7& \textbf{30.3}&   \textbf{37.9}&   \textbf{5.6}&    \textbf{1.6}&    95.5&    \textbf{5.9}&   \textbf{8.0} & \underline{11.9} \\
\bottomrule
\end{tabular}
\end{table*}

%% file: tables/cifar/jsd_vs_fractals.tex
\newcommand{\cmark}{\textcolor{green}{\ensuremath{\mathbf{\checkmark}}}}
\newcommand{\xmark}{\textcolor{red}{\ensuremath{\mathbf{\times}}}}
\newcommand{\csymbol}{\textcolor{orange}{$\bullet$} \textcolor{violet}{$\bullet$}}
\newcommand{\gsymbol}{\textcolor{black}{$\bullet$} \textcolor{gray}{$\bullet$} \textcolor{lightgray}{$\bullet$}}

\begin{table}[ht]
\centering
\caption{Comparison with experimentation settings of JSD loss with grayscale fractals on Cifar-100 dataset with WRN-40 model. We implemented their settings to the best of our knowledge. (\csymbol) represents the original fractals in grayscale while (\gsymbol) represents the grayscale fractals. $\circledast$ authors proposed settings. \textbf{Bold} is best and \underline{underline} is second best.} 
\label{tab:jsd_vs_fractals}
\begin{tabular}{lcccccc}
\toprule
\textbf{Model} & \textbf{JSD} & \textbf{Fractals} & \textbf{Classification} & \textbf{Robustness} & \textbf{Adversaries} & \textbf{Consistency} \\
& & & \textbf{Error($\downarrow$)} & \textbf{mCE($\downarrow$)} & \textbf{Error($\downarrow$)} & \textbf{mFR($\downarrow$)} \\
\midrule
Baseline  & \xmark     & \xmark      & 20.99 & 	51.03 &	97.04 & 11.02 \\ \midrule
PixMix$\circledast$    & \xmark      & \csymbol    & 20.36 & 35.54 & 92.55 & 6.16 \\
PixMix    & \xmark     & \gsymbol    & 20.56 &	35.01 & 97.49 & 6.54 \\
IPMix     & \xmark     & \csymbol     & 20.35 & 30.84 & 94.92 & 5.85  \\ 
IPMix     & \xmark     & \gsymbol     & 20.21 & 30.68 & 95.96 & 5.92  \\
LayerMix  & \xmark     & \csymbol   & 20.77 & 31.11 &	92.66 &	5.62 \\
LayerMix  & \xmark     & \gsymbol   & 20.70 & 30.34 &	95.49 &	5.55 \\ \midrule \midrule
PixMix    & \cmark     & \csymbol    & 18.46 &	34.04 & 86.67 & 5.91 \\
PixMix    & \cmark     & \gsymbol    & 18.63 &	32.79 & 94.74 & 5.98 \\ 
IPMix$\circledast$    & \cmark       & \csymbol     & 19.33 & 28.78 & 91.84 & \textbf{4.66}  \\
IPMix     & \cmark     & \gsymbol     & 19.13 & 28.29 & 91.93 & \underline{4.57}  \\ 
LayerMix & \cmark & \csymbol   & 18.73 & 29.82 &	86.63 &	5.49 \\
LayerMix & \cmark & \gsymbol   & 18.01 & 28.89 &	89.07 &	5.30 \\
\midrule \midrule
LayerMix (400 epochs) & \xmark & \csymbol & 19.41 & 	28.48 & 92.25 & 5.28 \\
LayerMix (400 epochs) & \xmark & \gsymbol & 18.77 & 	\underline{27.63} & 96.14 & 5.09 \\
LayerMix (400 epochs) & \cmark & \csymbol & \underline{18.36} & 	27.98 & \underline{81.41} & 5.32 \\
LayerMix (400 epochs) & \cmark & \gsymbol & \textbf{17.79} & 	\textbf{27.12} & \textbf{84.54} & 5.18 \\
\bottomrule
\end{tabular}
\end{table}

%% file: tables/cifar/mag_vs_br.tex
\begin{table}[ht]
\centering
\caption{Performance metrics for different combinations of corruption magnitude vs blending ratio on Cifar-100 dataset with WRN-40 model.}
\label{tab:mag_vs_blending}
\begin{tabular}{ccccccc}
\toprule
\textbf{Magnitude-} & \textbf{Classification} & \textbf{Robustness} & \textbf{Adversaries} & \textbf{Calibration} & \textbf{Consistency} \\
\textbf{Blending} & \textbf{Error($\downarrow$)} & \textbf{mCE($\downarrow$)} & \textbf{Error($\downarrow$)} & \textbf{RMS($\downarrow$)} & \textbf{mFR($\downarrow$)} \\
\midrule
5-3 & 20.51          & \textbf{30.24} & 95.29          & 6.25          & 5.49 \\
5-4 & 20.73          & 30.64          & 95.24          & 6.93          & 5.53 \\
5-5 & 20.62          & 31.07          & 94.37          & 6.84          & 5.65 \\ \midrule
6-3 & 20.96          & 30.55          & 95.49          & 6.00          & 5.66 \\
6-4 & 20.79          & 30.75          & \textbf{94.66} & 6.76          & 5.49 \\
6-5 & 20.49          & 30.96          & 94.88          & 7.17          & 5.48 \\ \midrule
7-3 & 20.80          & 30.49          & 95.67          & 6.33          & 5.51 \\
7-4 & 20.71          & 30.74          & 95.25          & 6.49          & 5.41 \\
7-5 & 20.47          & 30.97          & 94.81          & 6.94          & 5.37 \\\midrule
8-3 & 20.70          & 30.34          & 95.49          & \textbf{5.92} & 5.55 \\
8-4 & 20.72          & 30.79          & 94.92          & 6.50          & 5.45 \\
8-5 & \textbf{20.42} & 30.82          & 95.15          & 6.40          & \textbf{5.36} \\
\midrule
mean (std) & 20.66 ($\pm$0.15) & 30.70 ($\pm$0.37)& 95.10($\pm$0.37) & 6.54 ($\pm$0.37) & 5.49 ($\pm$0.09)\\ 
\bottomrule
\end{tabular}
\end{table}

%% file: tables/cifar/diff_models.tex
\begin{table}[ht]
\centering
\caption{Performance metrics for different model architectures for CIFAR-100. }
\label{tab:diff_models}
\begin{tabular}{c|c|r|ccccc}
\toprule
\textbf{Model (Params)} & \textbf{Epochs} & \textbf{Type} &\textbf{Classification} & \textbf{Robustness} & \textbf{Adversaries} & \textbf{Consistency} & \textbf{Calibration}  \\
& & & \textbf{Error($\downarrow$)} & \textbf{mCE($\downarrow$)} & \textbf{Error($\downarrow$)} & \textbf{mFR($\downarrow$)} & \textbf{RMS($\downarrow$)}\\
\midrule
\multirow{4}{6em}{WRN-40-4 (8.9M)} & \multirow{4}{*}{100} 
& baseline    & 20.99 & 51.03 & 97.04 & 11.02 & 14.24 \\ 
& & PixMix    & 20.36 & 35.54 & \textbf{92.41} & 6.16 & 7.00 \\  
& & IPMix     & \textbf{20.35} & 30.84 & 94.98 & 5.85 & 10.31 \\ 
& & LayerMix  & 20.70 & \textbf{30.34} & 95.49 & \textbf{5.55} & \textbf{5.92} \\
\hline \hline 
\multirow{4}{6em}{WRN-28-10 (36.5M)} & \multirow{4}{*}{100} 
& baseline   & 19.08 & 48.39 & 96.50 & 9.62 & 9.79  \\ 
& & PixMix   & 18.42 & 32.35 & \textbf{92.60} & 5.67 & 7.24 \\  
& & IPMix    & 18.81 & 28.58 & 94.84 & 5.32 & 9.21  \\ 
& & LayerMix & \textbf{18.05} & \textbf{27.41} & 95.24 & 4.94 & \textbf{5.83}  \\ 
\hline \hline 
\multirow{4}{6em}{ResNet-18 (11.2M)} & \multirow{4}{*}{200} 
& baseline    & 21.49 & 50.40 & 97.06 & 10.77 & 6.99 \\ 
& & PixMix    & \textbf{21.25} & 34.72 & \textbf{95.14} & 6.76 & 7.25 \\ 
& & IPMix     & 22.22 & 31.66 & 96.67 & 6.47 & 8.37 \\ 
& & LayerMix  & 21.37 &	\textbf{30.69} & 95.87 &\textbf{ 5.96} & \textbf{6.95} \\ 
\hline \hline 
\multirow{4}{6em}{ResNeXt-29 (6.9M)} & \multirow{4}{*}{200} 
& baseline    & 19.98 & 52.06 & 98.64 & 12.05 & \textbf{5.50} \\ %
& & PixMix    & \textbf{18.65} & 34.15 & 98.57 & 6.88 & 6.01 \\ %
& & IPMix     & 20.39 & 30.28 & \textbf{98.02} & 6.22 & 7.55 \\ 
& & LayerMix  & 18.94 & \textbf{28.75} & 98.71 &	\textbf{5.78} & 6.51 \\ %
\hline
\bottomrule
\end{tabular}
\end{table}

%% file: tables/imagenet/imagenet200.tex
\begin{table*}[ht]
\setlength\tabcolsep{5pt}
\centering
\caption{Full results for ImageNet-200. \textbf{Bold} is best, and \underline{underline} is second best.}
\label{tab:imagenet200}
\begin{tabular}{rcccccccccccc}
\toprule
 & \multicolumn{2}{c}{Accuracy}                      & \multicolumn{4}{c}{Robustness}        & \multicolumn{2}{c}{Consistency} & \multicolumn{4}{c}{Calibration}\\
 \cmidrule(lr){2-3}                                  \cmidrule(lr){4-7}                        \cmidrule(lr){8-9} \cmidrule(lr){10-13}
 & Clean@1 & Clean@5                               & C & $\overline{\text{C}}$ & R@1 & R@5 & \multicolumn{2}{c}{ImageNet-P} & Clean & C & $\overline{\text{C}}$ & R \\
 & \textcolor{gray}{Error} & \textcolor{gray}{Error} & \textcolor{gray}{mCE} & \textcolor{gray}{mCE} & \textcolor{gray}{Error} & \textcolor{gray}{Error} & \textcolor{gray}{mFR} & \textcolor{gray}{mT5D} &  \textcolor{gray}{RMS} & \textcolor{gray}{RMS} & \textcolor{gray}{RMS} & \textcolor{gray}{RMS}\\
\hline        %
 Baseline & 8.53 & 1.96 & 44.03 & 44.89 & 65.40 & 49.41 & 43.34 & 6.76 & \underline{2.60} & 11.71 & 15.04 & 23.16 \\
 Cutmix   & \textbf{7.81} & 1.87 & 45.26 & 44.49 & 66.46 & 51.05 & 47.54 & 7.45 & 4.88 & 10.51 & 12.90 & 21.45 \\
 Mixup    & \underline{7.87} & 1.72 & 37.60 & 37.47 & 62.24 & 46.31 & 43.27 & 6.80 & 3.96 & 7.08  & 11.07 & \underline{16.23} \\
 RE       & 8.21 & \textbf{1.65} & 44.01 & 44.48 & 66.19 & 50.40 & 45.12 & 7.00 & 3.38 & 12.39 & 14.88 & 24.26 \\
 AugMix   & 8.09 & 1.75 & 35.40 & 38.61 & 58.94 & 43.25 & 43.91 & 6.81 & 2.75 & 6.97  & 12.29 & 18.79 \\
 PixMix   & 7.97 & 1.86 & \underline{33.31} & \textbf{27.57} & \underline{57.64} & \underline{41.96} & 47.21 & 7.26 & 2.94 & \underline{5.16}  & \textbf{5.48} & \textbf{16.11} \\
 IPMix    & 7.93 & \underline{1.69} & 34.06 & 36.03 & 59.65 & 43.41 & \textbf{42.54} & \textbf{6.63} & 2.99 & 6.63  & 10.54 & 19.33 \\
 LayerMix & 8.05 & \underline{1.69} & \textbf{29.77} & \underline{29.26} & \textbf{57.27} & \textbf{41.46} & \underline{43.09} & \underline{6.67} & \textbf{2.46} & \textbf{4.72}  & \underline{6.44} & 16.63 \\
 \bottomrule%
\end{tabular}
\end{table*}

%% file: tables/imagenet/full_results.tex
\begin{table*}[ht]
\centering
\caption{Comparison on ImageNet-1K. \textbf{Bold} is best, and \underline{underline} is second best.
}\label{tab:imagenet_1k}
\begin{tabular}{rcccccccccc}
\toprule
 & \multicolumn{2}{c}{Accuracy} & \multicolumn{4}{c}{Robustness} & \multicolumn{4}{c}{Calibration}\\
 \cmidrule(lr){2-3} \cmidrule(lr){4-7} \cmidrule(lr){8-11}
 & Clean@1 & Clean@5 & C & $\overline{\text{C}}$ & R@1 & R@5 & Clean & C & $\overline{\text{C}}$ & R \\
 & \textcolor{gray}{Error} & \textcolor{gray}{Error} & \textcolor{gray}{mCE} & \textcolor{gray}{mCE} & \textcolor{gray}{Error} & \textcolor{gray}{Error} & \textcolor{gray}{RMS} & \textcolor{gray}{RMS} & \textcolor{gray}{RMS} & \textcolor{gray}{RMS}\\
\midrule%
baseline              & 23.84 &	7.12 & 60.19& 61.29	& 63.84& 47.12& 5.60& 11.91 & 20.73 & 19.71 \\ 	
PixMix                & 23.73 &	7.05 & 52.87& 47.82	& 59.65& 44.23& 4.43& 	5.90 &  6.68 &10.20 \\ 	
IPMix                 & 22.51 &	6.41 & 53.64& 54.28	& 60.94& 45.06& 3.66& 	6.80 & 11.06 &11.20 \\ 	
LayerMix(1m-4$\beta$) & 23.53 &	6.79 & 53.15& 49.38	& 60.14& 44.40& 4.06& 	5.83 &  8.21 &10.38 \\ 	
LayerMix(8m-3$\beta$) & 23.57 &	6.88 & 51.04& 48.05	& 59.59& 43.97& 4.76& 	5.68 &  7.04 &10.42 \\
\bottomrule
\end{tabular}
\end{table*}

%% file: tables/data.tex
\begin{table}[h]
    \centering
    \caption{Publically available data sources.}
    \label{tab:data_sources}
    \begin{tabular}{l|l}
    \toprule
       CIFAR-10/100 & \url{www.cs.toronto.edu/~kriz/cifar.html}\\
       CIFAR-10/100-C  & \url{github.com/hendrycks/robustness}\\
       CIFAR-10-$\overbar{\text{C}}$  & \url{github.com/facebookresearch/augmentation-corruption}\\
       CIFAR-100-$\overbar{\text{C}}$  & \url{github.com/ahmadmughees/layermix}\\
       CIFAR-10/100-P  & \url{github.com/hendrycks/robustness}\\
    \midrule
       ImageNet-1K  & \url{www.image-net.org}\\
       ImageNet-C  & \url{github.com/hendrycks/robustness}\\
       ImageNet-$\overbar{\text{C}}$  & \url{github.com/facebookresearch/augmentation-corruption}\\
       ImageNet-P  & \url{github.com/hendrycks/robustness}\\
       ImageNet-R  & \url{github.com/hendrycks/imagenet-r}\\
       ImageNet-200  & ImageNet-1k subset with the same classes as ImageNet-R\tablefootnote{list of classes is available at \url{github.com/ahmadmughees/layermix}}\\
       
    \bottomrule
    \end{tabular}
\end{table}

%% file: supplementry.tex
\appendices %
\section{More samples.} \label{sec:more_samples}
\begin{figure}[h]
    \centering
    \includegraphics[width=1\linewidth]{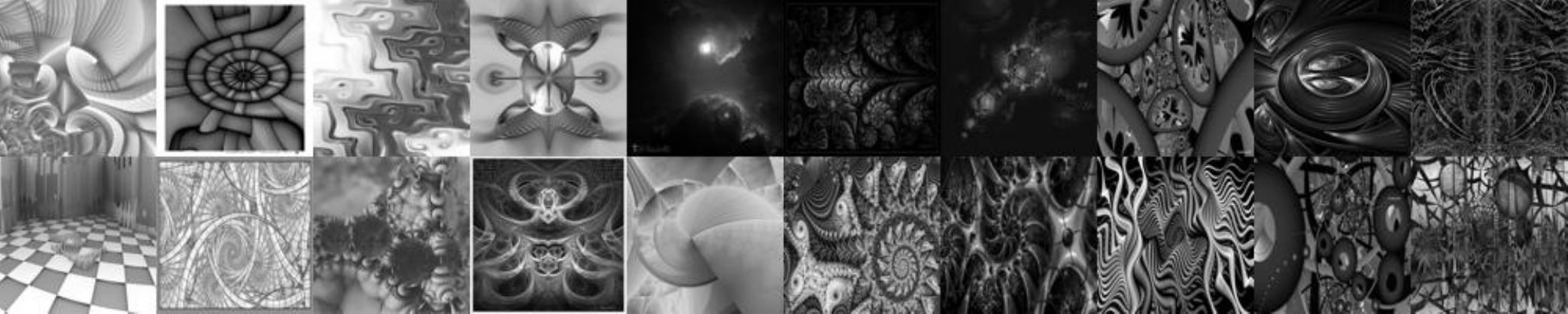}
    \caption{Samples of grayscale fractals used in LayerMix.}
    \label{fig:grayscale_fractals}
\end{figure}

\begin{figure}[h]
    \centering
    \includegraphics[width=0.85\linewidth]{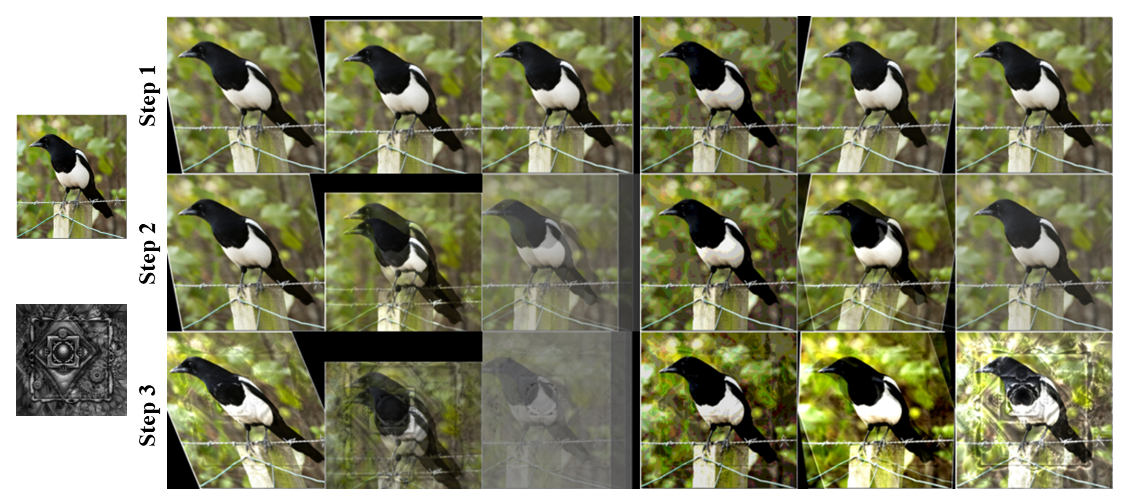}
    \caption{Samples generated from the 3 different layers of LayerMix.}
    \label{fig:grayscale_fractals}
\end{figure}

\begin{figure}[h]
    \centering
    \includegraphics[width=0.75\linewidth]{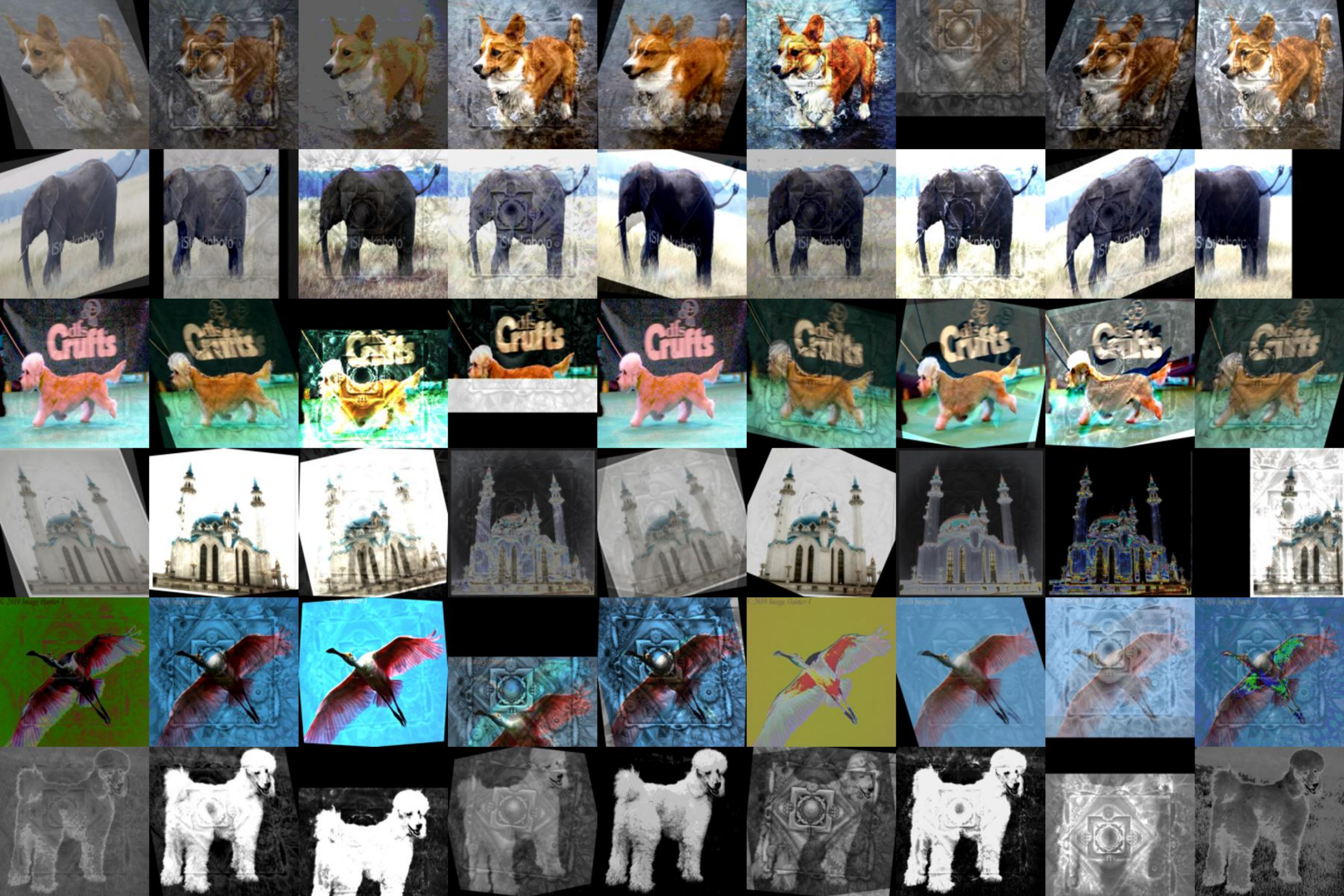}
    \caption{Samples generated from LayerMix.}
    \label{fig:grayscale_fractals}
\end{figure}

\begin{figure}[h]
    \centering
    \includegraphics[width=0.75\linewidth]{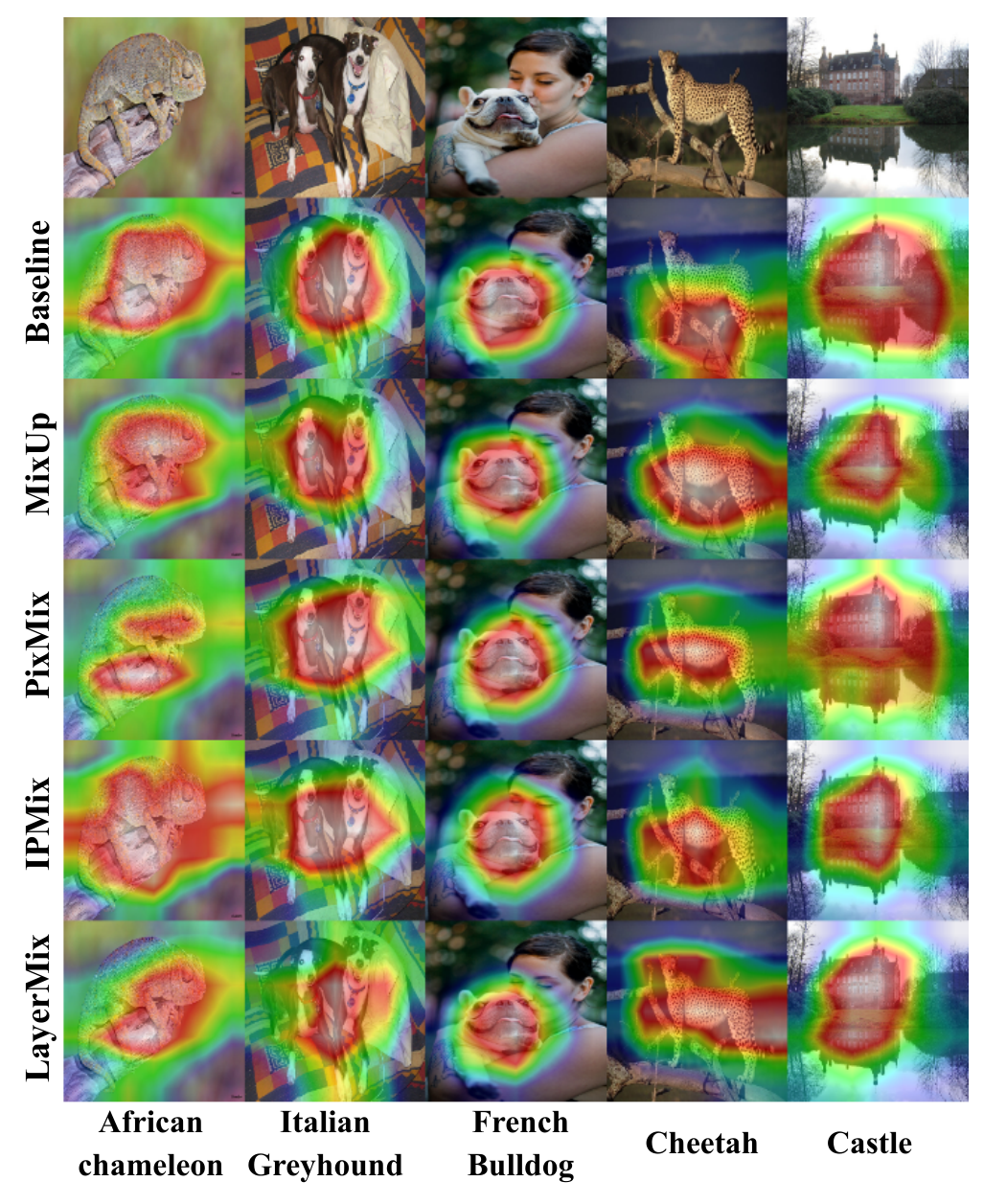}
    \caption{GradCam \cite{gradcam} class activation map visualizations from a ResNet-50 model across 5 randomly selected ImageNet samples for various augmentation pipelines.  The models evaluated in \cref{tab:imagenet200} were used for this comparison.}
    \label{fig:gradcam_viz}
\end{figure}

\section{Releted Work - Data Augmentation}
This section provides a comprehensive overview of our work's foundational components, focusing on data augmentation, complex image mixing strategies, and robustness evaluation benchmarks. 

\subsection{Image Augmentations}
These augmentations operate independently on a single data sample, introducing variations without relying on other samples from the dataset. Transformations of this type often generate samples that are closely clustered within the feature space. Examples include spatial transformations and affine manipulations~\cite{Lim2019FastA, Cubuk2019AutoAugmentLA, cubukRandAugmentPracticalAutomated2019}. Additionally, policy-driven augmentations have gained traction in recent years. AutoAugment~\cite{Cubuk2019AutoAugmentLA}, for instance, employs reinforcement learning to automatically discover optimal augmentation policies, while AdversarialAutoAugment~\cite{Zhang2019AdversarialA} generates adversarial examples to dynamically adjust augmentation strategies during training. 

Randomization-based approaches like RandomAugment~\cite{cubukRandAugmentPracticalAutomated2019} simplified the policy search by randomly applying augmentation operations in the sequence to generate unique synthetic samples. TrivialAugment~\cite{Mller2021TrivialAugmentTY} further extended it by applying only one augmentation operation in a pipeline. They sampled random corruption magnitude from a uniform distribution everytime, making them computationally efficient. Other notable methods include localized augmentation techniques such as CutOut~\cite{devriesImprovedRegularizationConvolutional2017} and RandomErase~\cite{Zhong2017RandomED} which randomly masks regions of an image to compel models to rely on broader spatial features, thereby improving generalization. 

\subsection{Image Blending}  
Image blending techniques leverage pixel-level mixing strategies to create diverse augmented samples. These methods often involve pixel-wise weighted averages to combine information from multiple images, enhancing robustness and generalization. MixUp~\cite{Zhang2018mixupBE} generates augmented samples by linearly interpolating between two randomly selected images and their corresponding labels, encouraging smoother decision boundaries for classification tasks. Manifold-MixUp~\cite{Verma2018ManifoldMB} performs similar interpolations within the hidden layers of a neural network, leading to improved accuracy by enforcing smooth transitions between feature representations.  

CutMix~\cite{Yun2019CutMixRS} introduces a patch-based approach, replacing a region of one image with a corresponding patch from another, while blending their labels proportionally to the patch size. This method enhances performance by encouraging the network to focus on multiple regions of interest. SaliencyMix~\cite{Uddin2020SaliencyMixAS} uses saliency maps to guide augmentation, replacing a square patch of the original image with the most salient regions from another image. This strategy ensures that the inserted regions carry highly informative features, thereby improving model learning.

AugMix~\cite{hendrycks*AugMixSimpleData2019} introduced a novel strategy by combining multiple transformations to produce diverse and robust augmented images, achieving state-of-the-art performance in robustness to corruptions and calibration tasks. Similarly, AutoMix~\cite{Liu2021AutoMixUT} employs a bi-level optimization framework to simultaneously improve the generation of mixed samples and the training of classifiers, achieving significant gains in robustness and accuracy.
TokenMix~\cite{Liu2022TokenMixRI}, designed for Vision Transformers, partitions images into multiple distinct regions at the token level and mixes them to exploit the unique properties of transformer architectures, resulting in better performance on vision tasks.
PuzzleMix~\cite{Kim2020PuzzleME}, SmoothMix\cite{leeSmoothMixSimpleEffective2020}, FMix\cite{harrisFMixEnhancingMixed2020},  LocalMixup~\cite{baenaLocalMixupInterpolation2024}, CutPaste~\cite{Li2021CutPasteSL} are few other models having similar topology. 

\subsection{Fractal-Based Augmentations}  
Fractals also known as non-natural images, are images with complex structures. They have intriguing properties that humans often rely on for perception. Such properties include structural characteristics of contours—such as orientation, length, and curvature—and junction types and angles derived from natural scene line drawings \cite{Walther2014NonaccidentalPU}. 

Fractal-based augmentation methods introduce structural complexity to the augmentation process by incorporating mathematically intricate patterns or synthetic structures to generate the labels-preserved samples that avoid the Manifold Intrusion \cite{guoMixUpLocallyLinear2019, baenaLocalMixupInterpolation2024} due to MixUp based approaches~\cite{Zhang2018mixupBE}.
PixMix~\cite{hendrycksPixMixDreamlikePictures2022} combines natural images with fractals and feature visualizations, which serve as highly diverse and complex augmentations to boost model robustness. They observed a significant accuracy boost for the out-of-domain benchmarks. IPMix~\cite{huangIPMixLabelpreservingData2024} adopts a three-stage approach, integrating images, patch, and pixel-based approach in label-preserving mixing strategies to generate robust training samples. Recent methods such as DiffuseMix~\cite{islamDiffuseMixLabelPreservingData2024} and GenMix~\cite{islamGenMixEffectiveData2024} blended the samples augmented through the Diffusion model with complex fractals and noticed the accuracy improvement.